\documentclass{ieeeaccess}
\usepackage{cite}
\usepackage{amsmath,amssymb,amsfonts}
\usepackage{algorithmic}
\usepackage{graphicx} 
\usepackage{subcaption}
\usepackage{array} 
\usepackage{ragged2e}
\usepackage{setspace}

\usepackage{multicol}
\usepackage{ragged2e}

\usepackage{afterpage}
\usepackage{booktabs}  
\usepackage{xcolor}
\usepackage{hyperref}
\usepackage{multirow}

\usepackage{placeins} 

\def\BibTeX{{\rm B\kern-.05em{\sc i\kern-.025em b}\kern-.08em
    T\kern-.1667em\lower.7ex\hbox{E}\kern-.125emX}}
\begin{document}
\history{Date of publication xxxx 00, 0000, date of current version xxxx 00, 0000.}
\doi{10.1109/ACCESS.2017.DOI}

\title{An evaluation of LLMs and Google Translate for translation of selected Indian languages via sentiment and semantic analyses}
\author{\uppercase{Rohitash Chandra}\authorrefmark{1}, \IEEEmembership{SM, IEEE},
\uppercase{Aryan Chaudhari\authorrefmark{2,3}, and Yeshwanth Rayavarapu}.\authorrefmark{1},
\IEEEmembership{Member, IEEE}}
\address[1]{Transitional Artificial Intelligence Research Group, School of Mathematics and Statistics, UNSW Sydney, Sydney, Australia}
\address[2]{Department of Civil Engineering, National Institute of Technology, Rourkela, India}
\address[3]{Centre for Artificial Intelligence and Innovation, Pingla Institute, Sydney, Australia}

\markboth
{Chandra \headeretal: IEEE Access}
{Chandra \headeretal: IEEE Access}

\corresp{Corresponding author: R. Chandra (e-mail: rohitash.chandra@unsw.edu.au)}


\begin{abstract}
 Large Language models (LLMs)  have been prominent for language translation, including low-resource languages.  There has been limited study on the assessment of the quality of translations generated by LLMs, including Gemini, GPT, and Google Translate. This study addresses this limitation by using semantic and sentiment analysis of selected LLMs for  Indian languages, including  Sanskrit, Telugu and Hindi. We select prominent texts (Bhagavad Gita, Tamas and Maha Prasthanam ) that have been well translated by experts and use LLMs to generate their translations into English, and provide a comparison with selected expert (human) translations.  
Our investigation revealed that while LLMs have made significant progress in translation accuracy, challenges remain in preserving sentiment and semantic integrity, especially in metaphorical and philosophical contexts for texts such as the Bhagavad Gita.  The sentiment analysis revealed that GPT models are better at preserving the sentiment polarity for the given texts when compared to human (expert) translation. The results revealed that GPT models are generally better at maintaining the sentiment and semantics  when compared to Google Translate. This study could help in the development
of accurate and culturally sensitive translation systems
for large language models. 
\end{abstract}


\titlepgskip=-15pt

\maketitle

\section{Introduction}
\label{sec:introduction}
Recent developments in the use of deep learning models \cite{sergeinirenburg_1993_progress} in natural language processing (NLP) \cite{cambria_2014_jumping}  had a significant impact on the field of machine translation (language translation) \cite{wang_2021_progress,li_2017_deep,wu_2019_deep}.
This led to the development of sophisticated models \cite{philippkoehn_2014_statistical} capable of understanding and translating text, such as Google Translate \cite{wu_2016_googles}.


Machine translation systems \cite{bennett_1985_the,riveratrigueros_2021_machine} refer to translation software systems that have achieved remarkable success in recent years due to multilingual pre-trained language models \cite{yao_2023_empowering}.  Large Language models (LLMs) \cite{zhao_2023_a,chang_2024_a,min_2023_recent} are pre-trained deep learning models that have emerged as a promising tool for translation systems and other NLP tasks such as multilingual machine translation \cite{aharoni_2019_massively}, text summarisation \cite{veniaminveselovsky_2023_artificial} and language modelling \cite{caliskan_2017_semantics}. However, the translation quality of these models, especially for low-resource languages  \cite{cahyawijaya_2024_llms}, often falls short of that achieved by models trained on higher-resource languages. Low resource languages refer to the less studied, resource-scarce, less computerised,  less commonly taught or low-density languages \cite{anilkumarsingh_2008_natural,cieri_2016_selection,tsvetkov_2017_opportunities}.   The scarcity of extensive monolingual or parallel corpora poses significant challenges to the development of accurate and culturally sensitive machine translation systems \cite{magueresse_lowresource, vaswani_2017_attention}. 

 Natural Language Generation (NLG)  \cite{schubert_2014_computational} enables computational devices to understand and write text that is indistinguishable from the work of humans across a vast array of languages.  Gemini is a multimodal LLM  developed by Google DeepMind and released in November 2023, which is a predecessor of PaLM (Pathways Language Model)   \cite{narang_2022_pathways}. Gemini is based on variants of Generalised Pre-trained Transformer (GPT) models \cite{floridi_2020_gpt3,liu_2023_gpt,kalyan_2023_a} that were originally released as Chat-GPT in November 2022 by OpenAI \cite{openai_2022_introducing}. Chat-GPT enables a conversation with user prompts and replies, which can also be used for language translation \cite{hendy_2023_how}.  Although prominent, GPT-based models are criticised for a lack of accountability and transparency \cite{hacker_2023_regulating,liesenfeld_2023_opening}, explainability \cite{zhao_2024_explainability}, and expertise in a specific field,  i.e. they have been called 'master of none' \cite{koco_2023_chatgpt}.                          Gemini and Chat-GPT have brought an entirely new era to machine translation \cite{wang_2019_learning,peng_2023_towards}, which is being validated by comparing with expert (human) translators \cite{lee_2023_artificial}. However, validating low-resource languages can be challenging due to constraints in the amount of data available    \cite{chi_2020_crosslingual}. The use of transfer learning\cite{zhang_2023_sentiment}, multilingual models\cite{chau_2021_specializing},  and hybrid systems \cite{ranathunga_neural} can improve the quality of translations.

Low-resource languages such as Telugu, Hindi and Sanskrit are often overlooked in the realm of language technology, including translation, speech recognition and NLP problems \cite{mohtashami_2023_learning}. 
These languages are characterised by their under-representation in digital datasets and a scarcity of linguistic resources. Hence, they present unique challenges for researchers and developers aiming to create effective language processing systems \cite{kumar_2022_annotated}. The complexity of these languages, coupled with the limited availability of resources, requires innovative approaches to translation and language processing.  Hendy et al. \cite{hendy_2023_how} reported that GPT models are better at translating high-resource models but are lacking when it comes to low-resource languages. Similarly, an approach by Sethi et al. \cite{sethi_2023_a} applied  GPT models for Sanskrit to Hindi translation, highlighting the potential of GPT models to enhance translation quality despite the scarcity of resources. These studies underscore the importance of developing innovative methods to address the challenges posed by low-resource languages.

A critical aspect of our study is the analysis of sentiment and semantic elements in the translated text. Sentiment analysis \cite{soujanyaporia_2018_multimodal, mustafa_2023_real} involves detecting the emotional tone expressed, typically via deep learning and LLMs.  
Chandra et al.\cite{shukla_an} presented an evaluation of Google Translate for Sanskrit to English translation via sentiment and semantic analysis. The study evaluated the translation quality of Sanskrit-English by Google Translate using sentiment and semantic analysis. Their study analysed the translation of the Bhagavad Gita by Google Translate and compared it with expert (human) translations, which motivates our study.

Semantic analysis \cite{goddard_2011_semantic, malik_2021_multimodal}, on the other hand, focuses on the meaning of words and phrases, ensuring that the translated text accurately conveys the intended meaning \cite{krommyda_2020_semantic}. Semantic analysis can check if the translation remains faithful to the original text in terms of meaning. Hence, we can compare LLM translations with expert  (human) translations using similarity scores such as the BLEU (Bilingual Evaluation Understudy) \cite{post_2018_a}.


In the past decades, there has been much focus on the analysis of the quality of machine translation, which is crucial for the advancement and practical implementation of translation applications. Gamon et al. \cite{gamon_2005_sentencelevel} developed an automatic evaluation process using BLEU and compared it with human translations via n-grams as an indirect indicator of translation quality. Hendy et al. \cite{hendy_2023_how} used GPT models for machine translation tasks by focusing on the zero-shot translation capabilities of GPT models, which are capable of performing well on multiple language pairs without any pre-training procedures or domain-specific customisation process \cite{mager_2020_GPTtoo}. This breakdown reveals their intrinsic language acquisition and expression abilities and the effect of making use of various prompting mechanisms on the way GPT models are translated. Zhang et al. \cite{zhang2023prompting} reviewed different types of prompts that trigger the most authentic and fluent translation produced by these language models to estimate the capacity of GPT models for document-level translation. Yang et al. \cite{yang_2020_towards}  examined the effects of  GPT and the traditional neural machine translation models, and reported that their synergy provides much higher quality. 

The extended research in this space encompasses recent machine translation methods and the rapid development of deep learning for NLP  \cite{lo_2020_extended}. Nevertheless, as machine translation more frequently attempts to cope with quality requirements, it is natural that formula-driven and context-less metrics such as BLEU  often do not reflect the intrinsic characteristics of the text. Furthermore, embedding-based metrics such as  YiSi-1 \cite{lo-2020-extended} and BERTScore \cite{zhang2020bertscore} have been demonstrated to be effective in giving back similar results with expert (human) translations.  There is a scarcity of research relating to the quality of the utterances generated by large language models. The design of efficient machine translation systems, especially those that work well in low-resource language settings, is a difficult project to do \cite{faheem_2024_improving}. Although NLP became more sophisticated with LLMs, the subtle nuances remain in the understanding of low-resource languages, including religious and philosophical texts \cite{petroanu_2023_tracing}. The neglected and under-resourced languages reflect the linguistic heritage, as well as the culture and identity of millions and billions of people around the world.

\textcolor{black}{In this study, we evaluate the potential of LLMs in translating selected low-resource Indian languages, using sentiment and semantic analysis. We use Google Translate, Gemini, and GPT-4o to translate prominent texts that are prominent in the region and translated by experts to provide a comparison with expert (human) translations. In order to validate the effectiveness of the translations, we employ different methodologies, including lexicon-based sentiment analysis and semantic analysis for comparison and evaluation of translation quality. We select   Sanskrit, Telugu and Hindi texts for our study and provide translated versions for further analysis via open-source code and data. Our research includes low-resource languages that are spoken by millions of people in the hope of preserving the beauty of human expression and looking for new ways to connect people from different cultures around the world.}


The rest of the paper is organised as follows. In Section 2, we present the background of LLMs and  NLP. In Section 3, we present the methodology used in the data extraction and processing for selected texts. We introduce a framework for sentiment and semantic analysis and present the experimental design.
In Section 4, we present the results of the sentiment analysis and semantic analysis for comparing the LLMs. Sections 5 and 6 provide a discussion of the results and the conclusion, respectively.

 \section{Background}
\subsection{Sentiment and Semantic Analysis}\label{lca}

Sentiment analysis and semantic analysis are two critical areas in  NLP that aim to understand and interpret human language. Sentiment analysis \cite{nasukawa_2003_sentiment}, also known as opinion mining, focuses on determining the sentiment or emotion expressed in a piece of text, such as whether the text conveys positive, negative, and neutral sentiments \cite{liu_2012_a}. Sentiment and semantic analysis become mutually complementary in a manner that allows for a complete interpretation of opinions, sentiments, entities, relations and the meaning of the surrounding context in the given text data \cite{daudert_2021_exploiting}. Implementing transfer learning and transformer-based language models has significantly improved the accuracy of implementing such systems \cite{zhao_2024_a}. Nonetheless, Misra et al. \cite{misra_2023_sarcasm} introduced a large-scale and high-quality dataset and framework for deep learning models that can reliably learn sarcastic cues from the text in an expressive manner. In the case of our study, a combination of sentiment and semantic analysis would be useful, as done in earlier studies \cite{zhang_2023_sentiment, chandra_2022_semantic}.
 
Sentiment analysis can be done at three levels: sentence level, document level and aspect level. At the sentence level, the document consisting of paragraphs is broken into sentences and sentiment analysis is done for every sentence  \cite{nandwani_2021_a}. Document-level sentiment analysis consists of the sentiment extraction from the entire document, intended for gaining global sentiment so that there will be no local unnecessary patterns\cite{medhat_2014_sentiment}. Differentiating the sharp distinction between the intricate internals of sentiments and the dependent words at this level demands greater insight. In the aspect-level sentiment analysis, emphasis is placed on opinions concerning the features or attributes that are expressed in the text \cite{wilson_2009_recognizing}.  LSTM models with attention mechanism \cite{park_2020_deep} addressed this difficulty by only distinguishing multiple aspects in a sentence, but eventually measuring the polarity of each aspect at the same time. 

The combination of sentiment analysis and semantic analysis techniques has become a powerful tool for mining comprehension of human language data at deeper levels \cite{jaminrahmanjim_2024_recent}. The main obligations of sentiment analysis lie in the classification of overall perceptions and feelings, yet semantic analysis concerns itself with contextual meaning, linguistic relationships and interpretation of the lexemes \cite{taboada_2016_sentiment}. 
The semantic analysis aims to derive meaning representations from text \cite{salloum_2020_a}, which encompasses a diverse set of techniques to uncover the underlying semantics, including named entity recognition, coreference resolution, semantic role labelling, and relationship extraction   \cite {khurana_2022_natural}. Chandra et al. \cite{chandra_2022_semantic} used a combination of sentiment and semantic analysis to compare different translations of the Bhagavad Gita (Sanskrit-English), which was later compared with Google Translate \cite{shukla2023evaluation}. These studies motivate their combination to be used for the comparison of LLMs for the translation of selected Indian languages.

\subsection{BERT}\label{lca}

The 
BERT  (Bidirectional Encoder Representations from Transformers) model  \cite{ddeepa_2021_bidirectional} provides a significant advancement in NLP. It employs a masked language modelling approach   \cite{lee_2021_combining} to analyse the whole text from both directions, providing more contextual understanding and better word embedding in NLP tasks such as sentiment analysis. The awareness of certain contexts is important for translation because the sense of a word may change upon emerging in different texts \cite{devlin_2018_bert}. In the case of machine translation using BERT, the context apprehension of BERT allows the model to accept literary devices such as idioms, and allegory  \cite{lee_2021_combining}. These usually confuse traditional machine translation systems that can't grasp or understand the context. This facilitates better translations that not only mean the beginning but also understand the tone and nuance \cite{ortizgarces_2024_optimizing}. Hence, BERT-based models have gained immense attention in NLP applications, including sentiment analysis \cite{acheampong2021transformer}. Prominent BERT-based implementations include  advanced models  (MPNet \cite{song2020mpnet}), hate speech analysis (HateBERT \cite{caselli2020hatebert}), and semantic analysis (Semantic-BERT\cite{zhang2020semantics}). Recently, Singh and Chandra released the  Hindubhobic-BERT model \cite{singh2025hp} for sentiment and abuse analysis of Hindubhobic text based on X (Twitter).  In our study, we will use BERT-based models for sentiment and semantic analysis as done in previous studies \cite{chandra_2022_semantic,shukla2023evaluation}

\section{Methodology}\label{cha}

\subsection{Data}\label{lca}


According to 2011 Indian Census,  Hindi is spoken by more than 43 \% of the Indian population as first and second languages \footnote{\url{https://web.archive.org/web/20180627091302/http://www.censusindia.gov.in/2011Census/C-16_25062018_NEW.pdf}}. Taking into account 2024 population projection by Worldometer \footnote{\url{https://www.worldometers.info/world-population/india-population/}}, we produce Table \ref{tab:language_population} that shows the percentage of Hindi, Telugu and Sanskrit speakers in 2011 and 2024. Although Hindi and Telugu feature large language populations, they would still be considered low-resource languages, due to the use of these languages in the education system,  resources available in language translation, and analyses of translations. 
\textcolor{black}{We remove the stop words and undesired characters for bigrams and  trigram analysis. In the case of translation  by LLMs,  we only remove the numbers featured in the respective texts.} Next, we present the selected texts for the three languages for our study.

\begin{table}[h]
    \centering
    \begin{tabular}{|l|l|l|c|}
        \hline
        Language & 2011 Census& Estimated 2024 & Growth (\%) \\
        \hline
        Hindi & 528,347,193 & 633,250,372 & 1.1979 \\
        Telugu & 81,127,740 & 97,421,412 & 1.2003 \\
        Sanskrit & 24,821 & 29,914 & 1.2040 \\
        \hline
    \end{tabular}
    \caption{Projected population of selected languages in India (2011 vs 2024).}
    \label{tab:language_population}
\end{table}


\subsubsection{Sanskrit: Bhagavad Gita}

The Bhagavad Gita \cite{shripurohitswami_2010_the} \cite{prabhupada1972bhagavad, jeste2008comparison} is one of the foundational texts of  Hinduism, encompassing practices, spirituality and philosophy.  The text features eighteen chapters and is divided into several parts, each of which addresses various philosophical topics, such as duty (dharma), action (karma),  spiritual way of life, and the nature of the self (Atman). The first chapter introduces the protagonists (Arjuna and Krishna)  and the problem setting of a philosophical nature based on an event from the Mahabharata, i.e. Arjuna's ethical dilemma to go or not to go to war for dharma (duty).

The original language of the Bhagavad Gita is Sanskrit, featuring 701 verses (sloka)   describing a conversation from the Mahabharata war in India. The Bhagavad Gita is a synthesis of various Hindu philosophies, including Vedanta, Yoga and Samkhya and discusses and synthesises the three dominant themes in Hinduism:  enlightenment-based renunciation,  dharma-based household life and devotion-based theism\cite{shripurohitswami_2010_the}. The Bhagavad Gita has been translated over the past two millennia and has been translated numerous times into hundreds of languages, with influence worldwide \cite{bayly2010india}. It has been studied with perspectives of leadership \cite{nayak2018effective}, psychology and psychotherapy \cite{bhatia2013bhagavad,jeste2008comparison}, and management \cite{mukherjee2017bhagavad}. As noted earlier, Shukla et al. \cite{shukla2023evaluation}   translated the Bhagavad Gita into English using Google Translate and compared different translations using sentiment analysis. Currently, Sanskrit is thriving in academia, given that the majority of the Hindu texts were originally written in Sanskrit,  but it is struggling as a language of everyday use and communication, as shown in Table \ref{tab:language_population}.

\subsubsection{Telugu: Maha Prasthanam}

Maha Prasthanam \cite{sastry1979sri}\footnote{\url{https://greaterTelugu.org/sri-sri-mahaprasthanam-pdf-book/}} is a Telugu-language anthology of poems by the noted literary writer Srirangam Srinivasarao, which features 18 chapters. It is this anthology that is considered an epic and a magnum opus in modern Indian poetry, which was published in 1950, and ever since, it has completely transformed the Telugu drama. The book is a poem collection which was written between 1930 and 1940, and when it was published, a milestone was made in Telugu literature. The title (Maha Prasthanam) is an amalgamation of two words, prasthanam meaning 'the way' and maha, the superlative for 'great'. This is an apt description of the depth of thematic and philosophical exploration in the poem. It has been translated into English by K. S. Sastry in 1979 \cite{sastry1979sri}, which made the work more accessible and thus makes it possible to analyse it to bridge the language gap between Telugu and English.

\subsubsection{Hindi: Tamas}

Tamas is a novel by Bhisham Sahni \cite{bhshmashan_2001_tamas} written in Hindi that describes the implications of the division of India in the aftermath of the partition of 1947 \cite{ahmed20021947}. More specifically,  in a solemn and heart-wrenching manner, it describes the 1947 riots. The novel is unique for its unusual narrative that does not depict the conflict through the life of a certain character alone; instead, it spreads a wide vision of conflict from the insides of various people of different races, political affiliations and religious beliefs. The story covers the riots as a result, which marked the British Partition of India. The division aimed to separate the major Hindu and Muslim populations into modern-day India and Pakistan, which ultimately resulted in bloody violence and displacement \cite{pandey2001remembering}. In 1988, Govind Nihalani adapted the book into a television series \footnote{Tamas: \url{https://www.imdb.com/title/tt14751340/}}      starring  Om Puri, which won multiple National Film Awards. The text features  21 chapters, and we will use a selected chapter for our study.

\subsection{Data processing}

We obtained prominent translations of the respective texts in English and converted each of the selected texts in a Printable Document Format (PDF). 
\textcolor{black}{In the case of the Sanskrit language (Bhagavad Gita), we used the entire text, and in the case of Telugu and Hindi, we used selected chapters that were openly available for usage.}
 We used the Natural Language Processing Toolkit (NLTK)  \footnote{\url{https://www.nltk.org/}} to remove undesired characters that can appear due to conversion.  Note that we obtain both the original and the translation of the respective texts, and the original text is written in Devanagari script in the case of Sanskrit (Bhagavad Gita) and Hindi (Tamas) and the Telugu (Maha Prasthanam), respectively. The Telugu script features  54 letters - 16 vowels, 3 vowel modifiers, and 35 consonants. The Devanagari script is one of the official scripts of  India and Nepal, composed of 48 primary characters, including 14 vowels and 34 consonants \cite{daniels1996world}.

The process for the sentiment and semantic analysis of the chosen books, i.e. Bhagavad Gita, Maha Prasthanam and Tamas, involves several stages. Tokenisation refers to the process of breaking down a piece of text, such as a  sentence or a paragraph,  into individual words or “tokens.”   Therefore, we begin with tokenisation, text cleaning and pre-processing, which marks the initial phase of data extraction and processing. 
Then they were tokenised by verses for the Bhagavad Gita and by poems for the Maha Prasthanam. In the case of Tamas, we focused on sentiments classified based on small paragraphs.
Afterwards, the tokenised text was translated using different machine translation models (LLMs). 

\subsection{Neural translation models}\label{lca}
\subsubsection{Google Translate}

Google Translate \cite{google_2006_google}  marked a significant advancement in machine translation using Google Neural Machine Translation (NMT) \cite{johnson2017google}, leveraging statistical methods and later incorporating deep learning models \cite{wu2016google}. Google Translate supports translation between over 100 languages, ranging from widely spoken languages such as English, Spanish, and Chinese to less common ones such as Icelandic and Maori\cite{groves2015}. Some studies have reviewed the effectiveness of Google Translate, including translations in medical communication \cite{patil2014} and Sanskrit to English translation \cite{shukla2023evaluation}.  Tsai et al. \cite{Tsaishu2019} demonstrated that Google Translate English texts presented several components of significantly higher writing quality with fewer mistakes in spelling and grammar and fewer errors per word than its counterparts \cite{Tsaishu2019}.
Prates et al. \cite{Prates2020} studied gender biases and showed that Google Translate yields male defaults much more frequently than what would be indicated from demographic data alone. These studies motivate the comparison of Google  Translate with emerging LLMs, including GPT and Gemini.

\subsubsection{GPT models}\label{lca} 
ChatGPT  \footnote{ChatGPT: https://openai.com/chatgpt}   is a highly sophisticated LLM    \cite{wu2023brief} that revolutionised NLP  with chatbot applications that include query answering systems, language translation, and tasks such as text summarisation, literature review and essay writing, along with poems, prose, and songs \cite{fui2023generative}. GPT models use large amounts of data that include the Wikipedia corpus \cite{rogers_2023_chatgpt}. However, contrary to their ability to address various real-world tasks, the accuracy can be varied depending on how generic or discerning the input is leading to adequate or off-topic responses \cite{labruna2023unraveling,koco_2023_chatgpt}.    ChatGPT has been constantly updating its features to support more languages and has been widely adopted by people from different regions of the world. Furthermore, GPT-4o is the advanced version of ChatGPT which is accessible through \textit{ChatGPT Plus}, which has a monthly subscription fee and has access to information up to December 2023, and can also search for real-time data.  GPT-4o is multimodal, i.e. it can utilise data that includes text, speech and images. ChatGPT and GPT-4o leverage supervised learning and reinforcement learning from human feedback. ChatGPT supports more than 50 languages which are including English, French, German, Spanish, Chinese, Japanese and Arabic, as well as other more doable languages \cite{lai_2023_chatgpt}.   We will evaluate both Chat-GPT (GPT-3.5) and GPT-4o for translation of selected languages in our study.

Gemini has been developed by Google  \cite{_2021_Gemini} as a GPT-based LLM  that  \cite{syedaselinaakter_2023_an} uses English as the main language. The latest update (Gemini 3) incorporates more language versions: Italian, Portuguese and Turkish, ensuring the service availability in over 70 countries.
Gemini improved on the deficiencies of GPT-3 in terms of contextual understanding using a much larger text corpus and yielding more lengthy and well-structured narratives. However, the performance of  Gemini is similar to ChatGPT, and more research is needed to compare their performance for various tasks.  In our study, we will specifically assess the potential of Gemini in the translation of Indian texts, when compared to other LLMs and Google Translate.

 GPT-based models have been criticised for a lack of accountability, transparency and expandability \cite{benneh2023}. They can provide plausible, correct answers, which at times turn out to be incorrect with expert validation. They also have a bias due to the training data type and source. Although these models have been recently introduced, studies about their strengths and weaknesses are being reported, eg. Lee et al. \cite{lee2023gemini} reported that Gemini defeated GPT-4o in educational settings, where the findings suggest GPT-4o's superior capability in handling complex multimodal educational tasks.  Furthermore, Borji and Mohammadian  \cite{borji2023} reported that GPT-4 emerged as the top-performing chatbot across almost all categories. These categories include reasoning, logic, facts, coding, bias, language, and humour. Gemini is less correlated with other models, while ChatGPT and GPT-4 are highly correlated in terms of their responses.

\subsection{Semantic and Sentiment Analysis}


 MPNet \cite{ji_2020_leveraging} is an advanced  BERT model that is capable of semantic analysis while generating sentence embeddings.  MPNet employs permuted language modelling, (unlike Masked Language Modelling in BERT) and uses auxiliary position information as input and thus reducing the position discrepancy, enabling the model to utilise the full sentence. Hence, the sentences are assigned vector representations that capture the semantic meaning of the sentences \cite{yu_2019_masked}. These can be then compared to understand the similitude or dissimilarity between sentences. Therefore, MPNet has been specifically helpful for semantic search, which finds sentences that have a similar semantic meaning about a fixed query \cite{choi_2021_evaluation}. 
 
 KeyBERT  \cite{koloski_2022_out} is a keyword extraction model that makes use of BERT-based embedding to locate the most significant keywords in a text. It can determine the main issues covered in a text by extracting keywords that are the most representative of the content. Thus, KeyBERT can be trained on keywords that are relevant to particular subjects or themes, which enhances the tool`s ability to outline the main concepts in texts. This can be helpful when searching for common ideas in the translations and comparing them to the original texts \cite{nadim_2023_a}. In our study, we preprocess and tokenise the text using BERT for sentiment analysis of the text. We will employ MPNet for semantic analysis to compare the meaning of sentences across languages and  utilise KeyBERT for keyword extraction to identify the main topics in the text \cite{R2023}.

\subsection{Framework}

\begin{figure*}[htbp!]
        \centering
        \includegraphics[width=1\linewidth]{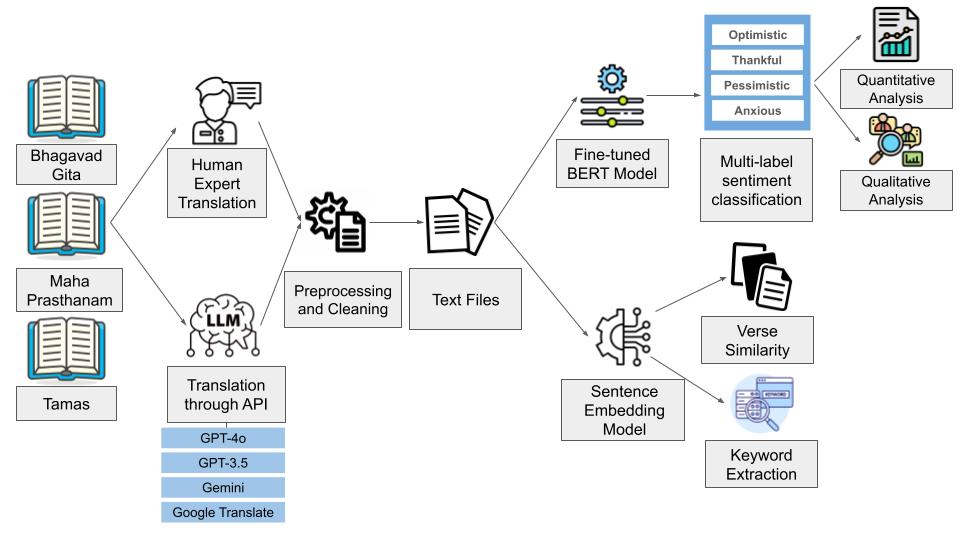}
        \caption{Sentiment and Semantic analysis framework for comparison of translations by LLMs (Google Translate, GPT-4o, GPT-3.5, and Gemini).}
        \label{fig:framework}
\end{figure*}

We present the framework in Figure \ref{fig:framework} for evaluating Google Translate, Gemini and GPT models for the translation of selected Indian languages, including  Telugu, Hindi and Sanskrit into English. We compare the translation of the texts with human-expert translation using semantic and sentiment analysis.  Each step in our framework plays a crucial role in analysing and comparing the translations using sentiment and semantic meaning.

We first begin (Step 1)  by processing the text in the electronic copies of the Bhagavad Gita (Sanskrit), Tamas (Hindi) and Maha Prasthanam (Telugu) obtained in their original script (Devanagari and Telugu).  We split the texts into slokas (verses) in the case of Bhagavad Gita and sentences in the case of Tamas and Maha Prasthanam ordered by chapters after preprocessing. We also obtain expert translations of the texts, as shown in Table \ref{tab:translationtable}.

\begin{table*}[h!]
    \centering
    \renewcommand{\arraystretch}{1.2} 
    \small 
    \begin{tabular}{|l|l|l|l|l|l|l|}
        \hline
        \textbf{Language} & \textbf{Text} & \textbf{Author} & \textbf{Publication Year} &  \textbf{Translator} & \textbf{Translation Year (Publisher)} \\
        \hline
        Sanskrit & Bhagavad Gita & -- & 300-500 BCE &Shri Purohit Swami & 1935 \\
        Telugu & Maha Prasthanam & 	Srirangam Srinivasarao (Sri Sri)&	1950 & Sri Sri  & 1970\\
        Hindi & Tamas & Bhisham Sahni  & 1974 &Bhisham Sahni  & 2001 (Penguin)\\
        \hline
    \end{tabular}
    \caption{Details about authorship, date of publication and translation for the selected texts. Note that the original author of the Bhagavad Gita is unknown, and its actual date of composition is difficult to access since it has been orally transmitted for centuries/millennia. }
    \label{tab:translationtable}
    
\end{table*}

In Step 2, we translate each of the given texts from their source language into English using the API (Application Programming Interface) of the LLMs (GPT and Gemini) and Google Translate.
We store the translated text in a CSV file and publish it via our GitHub repo \footnote{\url{https://github.com/sydney-machine-learning/GPTvsGoogleTranslate}}.

In Step 3, we provide translation vocabulary analysis using trigram analysis and compare the translations of the respective texts obtained by the LLMs and Google Translate.  We tokenise the text for sentiment analysis using the BERT model \cite{liu_2019_roberta}. We refine the BERT model using SenWave data \cite{yang2020senwave} for multi-label sentiment classification that can detect multiple emotions for a single text, which can be used for a deeper understanding of translation sentiments.   \textcolor{black}{The main motivation of using BERT-based SenWave model for sentiment analysis is due to its multi-label classification nature,  where more than one sentiment can be detected in a sentence. Moreover, it has been used effectively in previous research \cite{chandra_2022_semantic,shukla2023evaluation}  for sentiment analysis in the scope of machine translation. However, we note the limitation that the SenWave dataset is X (Twitter) dataset, while the applications in this study consists of philosophical and sacred texts and novellas.}

In Step 4, we use the MPNet model for semantic analysis and obtain a cosine similarity score between the expert translations and the translations. We then apply keyword extraction (Step 5) using KeyBERT, which will extract the keywords from the text, which can be analysed to know which keywords are dominant in various translations, and we also extract the trigrams using KeyBERT.

 In Step 6, we take the selected translated text from Step 2 for comparison with expert translation using qualitative analysis. Finally, we apply further visualisation and analysis  by focusing on specific aspects of the translations, such as the sentiment polarity over time for Arjuna and Lord Krishna of the Bhagavad Gita.

\section{Results}
\subsection{Data Analysis }
 We first present the trigram analysis comparing a human expert with machine translations that include Google Translate and LLMs. We removed stop words in the data preprocessing step, for trigram analysis, to provide insights into the difference in the vocabulary by the different LLMs for the translations. The trigram frequency analysis reveals key differences in how various models and human experts translate and interpret the text. 
 
 As seen in Table~\ref{tab:tri-sanskrit}, Google Translate strongly prefers a translation using theological terms, with high repetition of phrases ('supreme', 'personality', 'godhead'), suggesting a literal but somewhat redundant approach. Gemini focuses heavily on divine references, frequently using words such as  'krishna' and 'lord,' which portrays a devotional tone. GPT-3.5 and GPT-4o provide a more balanced interpretation, incorporating philosophical    ('goodness', 'passion', 'ignorance') and duty-related concepts ('perform', 'prescribed', 'duties') that are associated with the philosophy of karma, which is central to the Bhagavad Gita. In contrast, human expert annotations emphasise dialogue and context, using phrases such as  ('lord', 'shri', 'krishna') and ('arjuna', 'asked', 'lord'), which reflect the narrative style.

The trigram analysis for translations of Maha Prasthanam from Telugu-English (Table~\ref{tab:tri-telugu}) reveals distinct patterns across different models and human expert translations. Google Translate heavily favours repetition, particularly with phrases ('forward', 'forward', 'forward'), indicating a mechanical translation approach. GPT-3.5 and GPT-4o provide more varied outputs, with a mix of movement-related words ('step', 'forward', 'step') and ('come', 'forward', 'come'), though some trigrams appear redundant. Gemini emphasises directional movement, frequently using 'push' and 'forward,' which aligns with the theme of progress. In contrast, the human expert translation preserves poetic and philosophical depth, using phrases  ('forward', 'march', 'onward') and ('manifold', 'divine', 'lights'), which better capture the literary essence. This comparison highlights that while the different models can extract structure and meaning, the human expert translation retains the deeper poetic and contextual nuances of the original text.

In the case of Tamas translation from Hindi-English (Table~\ref{tab:tri-hindi}) GPT-3.5 and GPT-4o generate a mix of descriptive and action-based trigrams, such as ("pigs", "roam", "around") and ("walking", "along", "wall"), capturing movement and scene details. Google Translate produces a mix of mechanically translated phrases, often repeating phrases ("opened", "door", "closet"), showing a more direct and structured translation. We observe the common trigram ("flickering", "light", "lamp")  in GPT-3.5, GPT-4o, and Gemini, indicating a shared interpretation of the text's imagery. However, in the Google Translate output, the phrases ("lamp", "started", and "blinking") appeared instead, reflecting a more literal translation from the original Hindi text.

\begin{table*}[h!]
    \centering
    \small
    \renewcommand{\arraystretch}{1.2}
    \resizebox{\textwidth}{!}{ 
    \begin{tabular}{|c|c|c|c|c|c|}
        \hline
        \multicolumn{2}{|c|}{GPT-3.5} & \multicolumn{2}{c|}{GPT-4o} & \multicolumn{2}{c|}{Gemini} \\
        \hline
        Trigram & Freq & Trigram & Freq & Trigram & Freq \\
        \hline
        ('supreme', 'personality', 'godhead') & 13 & ('qualities', 'born', 'nature') & 5 & ('blessed', 'lord', 'krishna') & 63 \\
        ('mighty', 'armed', 'arjuna') & 7 & ('field', 'knower', 'field') & 4 & ('lord', 'krishna', 'person') & 6 \\
        ('goodness', 'passion', 'ignorance') & 4 & ('nature', 'goodness', 'sattvic') & 4 & ('lord', 'krishna', 'partha') & 5 \\
        ('perform', 'prescribed', 'duties') & 4 & ('attains', 'supreme', 'goal') & 3 & ('sacrifice', 'charity', 'austerity') & 5 \\
        ('duty', 'without', 'attachment') & 3 & ('cold', 'heat', 'pleasure') & 3 & ('attains', 'supreme', 'state') & 4 \\
        \hline
    \end{tabular}
    }

    \vspace{0.5cm} 

    \resizebox{0.7\textwidth}{!}{ 
    \begin{tabular}{|c|c|c|c|}
        \hline
        \multicolumn{2}{|c|}{Google Translate} & \multicolumn{2}{c|}{Human Expert} \\
        \hline
        Trigram & Freq & Trigram & Freq \\
        \hline
        ('supreme', 'personality', 'godhead') & 157 & ('lord', 'shri', 'krishna') & 38 \\
        ('personality', 'godhead', 'supreme') & 36 & ('shri', 'krishna', 'replied') & 17 \\
        ('godhead', 'supreme', 'personality') & 35 & ('arjuna', 'asked', 'lord') & 8 \\
        ('modes', 'material', 'nature') & 19 & ('krishna', 'replied', 'beloved') & 3 \\
        ('supreme', 'absolute', 'truth') & 15 & ('purity', 'passion', 'ignorance') & 3 \\
        \hline
    \end{tabular}
    }
    \caption{Trigrams Analysis for Bhagavad Gita translation from Sanskrit to English with LLMs, Google Translate and Human Expert.}
    \label{tab:tri-sanskrit}
\end{table*}

\begin{table*}[h]
    \centering
    \small
    \renewcommand{\arraystretch}{1.2}
    \resizebox{\textwidth}{!}{ 
    \begin{tabular}{|c|c|c|c|c|c|}
        \hline
        \multicolumn{2}{|c|}{GPT-3.5} & \multicolumn{2}{c|}{GPT-4o} & \multicolumn{2}{c|}{Gemini} \\
        \hline
        Trigram & Freq & Trigram & Freq & Trigram & Freq \\
        \hline
        ('step', 'forward', 'step') & 8 & ('another', 'world', 'another') & 4 & ('forward', 'push', 'forward') & 8 \\
        ('forward', 'step', 'backward') & 7 & ('world', 'another', 'world') & 4 & ('push', 'forward', 'push') & 6 \\
        ('another', 'world', 'another') & 4 & ('come', 'forward', 'come') & 4 & ('forward', 'move', 'forward') & 6 \\
        ('observations', 'many', 'directions') & 4 & ('hara', 'hara', 'hara') & 3 & ('another', 'world', 'another') & 4 \\
        ('world', 'another', 'world') & 4 & ('come', 'forward', 'push') & 3 & ('lamp', 'looks', 'directions') & 4 \\
        \hline
    \end{tabular}
    }

    \vspace{0.5cm} 

    \resizebox{0.7\textwidth}{!}{ 
    \begin{tabular}{|c|c|c|c|}
        \hline
        \multicolumn{2}{|c|}{Google Translate} & \multicolumn{2}{c|}{Human Expert} \\
        \hline
        Trigram & Freq & Trigram & Freq \\
        \hline
        ('forward', 'forward', 'forward') & 35 & ('forward', 'march', 'onward') & 8 \\
        ('hara', 'hara', 'hara') & 6 & ('march', 'forward', 'march') & 8 \\
        ('another', 'world', 'another') & 4 & ('another', 'world', 'another') & 4 \\
        ('looks', 'many', 'directions') & 4 & ('manifold', 'divine', 'lights') & 4 \\
        ('world', 'another', 'world') & 4 & ('march', 'onward', 'upward') & 4 \\
        \hline
    \end{tabular}
    }
    \caption{Trigrams Analysis for Maha Prasthanam translation from Telugu to English with LLMs, Google Translate and Human Expert.}
    \label{tab:tri-telugu}
\end{table*}

\begin{table*}[h]
    \centering
    \small 
    \renewcommand{\arraystretch}{1.2}
    \resizebox{\textwidth}{!}{ 
    \begin{tabular}{|c|c|c|c|c|c|}
        \hline
        \multicolumn{2}{|c|}{GPT-3.5} & \multicolumn{2}{c|}{GPT-4o} & \multicolumn{2}{c|}{Gemini} \\
        \hline
        Trigram & Freq & Trigram & Freq & Trigram & Freq \\
        \hline
        ('pigs', 'roam', 'around') & 2 & ('five', 'rupee', 'note') & 2 & ('lamp', 'niche', 'flickered') & 2 \\
        ('quickly', 'opened', 'door') & 2 & ('walking', 'along', 'wall') & 2 & ('eyes', 'still', 'fixed') & 2 \\
        ('roam', 'around', 'catch') & 2 & ('felt', 'like', 'crying') & 2 & ('flickering', 'light', 'lamp') & 2 \\
        ('five', 'rupee', 'note') & 2 & ('right', 'hand', 'wall') & 2 & ('walk', 'along', 'wall') & 2 \\
        ('flickering', 'light', 'lamp') & 2 & ('flickering', 'light', 'lamp') & 2 & ('sometimes', 'walk', 'along') & 2 \\
        \hline
    \end{tabular}
    }

    \vspace{0.5cm} 

    \resizebox{0.7\textwidth}{!}{ 
    \begin{tabular}{|c|c|c|c|}
        \hline
        \multicolumn{2}{|c|}{Google Translate} & \multicolumn{2}{c|}{Human Expert} \\
        \hline
        Trigram & Freq & Trigram & Freq \\
        \hline
        ('walking', 'along', 'wall') & 3 & ('thin', 'cane', 'stick') & 3 \\
        ('open', 'door', 'closet') & 2 & ('five', 'rupee', 'note') & 2 \\
        ('opened', 'door', 'closet') & 2 & ('flame', 'clay', 'lamp') & 2 \\
        ('chest', 'still', 'next') & 2 & ('light', 'clay', 'lamp') & 2 \\
        ('lamp', 'started', 'blinking') & 2 & ('little', 'tail', 'curling') & 2 \\
        \hline
    \end{tabular}
    }
    \caption{Trigrams Analysis for Tamas translation from Hindi to English with LLMs, Google Translate and Human Expert.}
    \label{tab:tri-hindi}
\end{table*}

\subsection{Sentiment Analysis}\label{lca}
\begin{table*}[h!]
    \centering
    \small
    \begin{tabular}{@{}lcccccccccc@{}}
        \toprule
        \textbf{Models} & \textbf{Optimistic} & \textbf{Thankful} & \textbf{Empathetic} & \textbf{Pessimistic} & \textbf{Anxious} & \textbf{Sad} & \textbf{Annoyed} & \textbf{Denial} & \textbf{Humour}  \\ 
        \midrule
        GPT-3.5          & 223                & 4                 & 4                   & 213                 & 66              & 372         & 235            & 152           & 10             \\ 
        GPT-4o            & 225                & 3                 & 5                   & 230                 & 59              & 375         & 225            & 161           & 7              \\ 
        Gemini           & 266                & 3                 & 3                   & 208                 & 75              & 322         & 194            & 133           & 9              \\ 
        Google Translate & 293                & 7                 & 8                   & 274                 & 71              & 332         & 192            & 151           & 5              \\ 
        Human Expert     & 223                & 3                 & 4                   & 228                 & 49              & 352         & 245            & 166           & 2              \\ 
        \bottomrule
    \end{tabular}
    \caption{Sentiments Counts(Sanskrit)}
    \label{tab:sentiments-sanskrit}
\end{table*}

\begin{table*}[h!]
    \centering
    \small
    \begin{tabular}{@{}lcccccccccc@{}}
        \toprule
        \textbf{Models} & \textbf{Optimistic} & \textbf{Thankful} & \textbf{Empathetic} & \textbf{Pessimistic} & \textbf{Anxious} & \textbf{Sad} & \textbf{Annoyed} & \textbf{Denial}  \\ 
        \midrule
        GPT-3.5          & 17                & 0                 & 0                   & 29                 & 10              & 15         & 9            & 0           \\ 
        GPT-4o            & 14               & 0                 & 0                   & 35                 & 7              & 18         & 9            & 0           \\ 
        Gemini           & 15               & 0                 & 0                   & 34                 & 10              & 14         & 4            & 0          \\ 
        Google Translate & 10                & 0                 & 3                   & 29                 & 8              & 17         & 14            & 2           \\ 
        Human Expert     & 19                & 0                 & 4                   & 33                 & 8              & 13         & 10            & 0           \\ 
        \bottomrule
    \end{tabular}
    \caption{Sentiments Counts(Telugu)}
    \label{tab:sentiments-telugu}
\end{table*}

\begin{table*}[h!]
    \centering
    \small
    \begin{tabular}{@{}lcccccccccc@{}}
        \toprule
        \textbf{Models} & \textbf{Optimistic} & \textbf{Thankful} & \textbf{Empathetic} & \textbf{Pessimistic} & \textbf{Anxious} & \textbf{Sad} & \textbf{Annoyed} & \textbf{Denial}  \\ 
        \midrule
        GPT-3.5          & 7                & 0                 & 1                   & 13                 & 15              & 18         & 24            & 4           \\ 
        GPT-4o            & 5               & 0                 & 0                   & 16                 & 20              & 20         & 26            & 7           \\ 
        Gemini           & 9               & 1                 & 0                   & 18                 & 15              & 18         & 23           & 5          \\ 
        Google Translate & 4                & 0                 & 0                  & 22                 & 17              & 22         & 30            & 3           \\ 
        Human Expert     & 6                & 0                 & 2                  & 26                 & 14              & 18         & 27            & 3           \\ 
        \bottomrule
    \end{tabular}
    \caption{Sentiments Counts(Hindi)}
    \label{tab:sentiments-hindi}
\end{table*}

\begin{figure}[htbp!]
        \centering
        \includegraphics[width=1\linewidth]{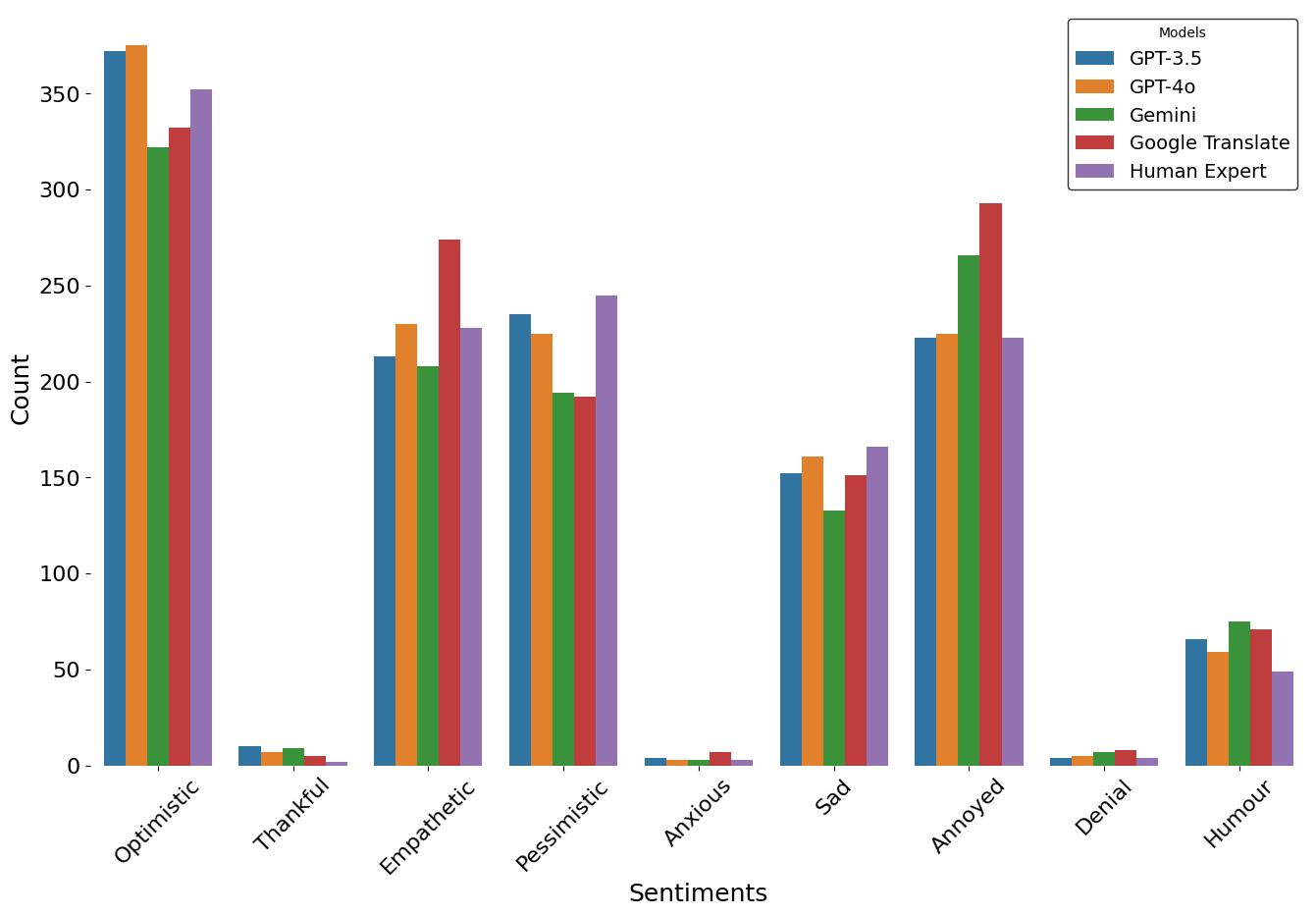}
        \caption{Sentiment Analysis from Google Translate, LLMs and Human Expert  translations (Sanskrit-English).}  
        \label{fig:Sanskrit}
    \end{figure}

We further present results from sentiment analysis of translations by LLMs (Gemini and GPT models)  for Sanskrit, Hindi and Telugu and compare the sentiments with human-expert translation. In this way, we can compare whether the different translations express similar sentiments along with the varied priority of the sentiments. In the case of the Bhagavad Gita translations from Sanskrit-English (Figure \ref{fig:Sanskrit}), we find that the respective models are visibly different in capturing various sentiments. We note that the "joking" sentiment refers to humour and this was detected by human expert translators such as Mahatma Gandhi, Sri Purohit Swami and Eknath Easwaren in a study by Chandra and Kulkarni \cite{chandra_2022_semantic}. Google Translate produced the highest number of "Optimistic" (293) and "Empathetic" (7) sentiments for Sanskrit-English translation (Figure \ref{fig:Sanskrit}), suggesting that its translation is more positively expressed sentiments than others. However, Google Translate also generated the highest number of "Pessimistic" (274) sentiments, indicating inconsistency in emotional tone. GPT-4o and the Human Expert had similar distributions across most sentiments, suggesting that GPT-4o’s translation is more aligned with human understanding. GPT-3.5 and GPT-4o had comparable sentiment counts, with GPT-4o slightly increasing the presence of emotions such as "Pessimistic" and "Denial."

Furthermore, Figure \ref{fig:Sanskrit} shows that Gemini's translations resulted in relatively lower "Sad" and "Annoyed" counts compared to GPT models and the Human Expert. However, it produced more "Anxious" (75) sentiments than others, indicating a potential shift in tone during translation. Interestingly, "Humour" was the least detected sentiment across all models, with Google Translate and the Human Expert showing the lowest counts (5 and 2, respectively). This suggests that humour is rare in the Bhagavad Gita's text. Overall, the variations in sentiment distributions highlight the differences in how each model interprets and translates Sanskrit to English. Although GPT-4o aligns closely with human translations, Google Translate shows extreme sentiment counts in terms of optimism and pessimism, and Gemini leans toward anxious expressions.

\begin{figure}[htbp!]
        \centering
        \includegraphics[width=1.05\linewidth]{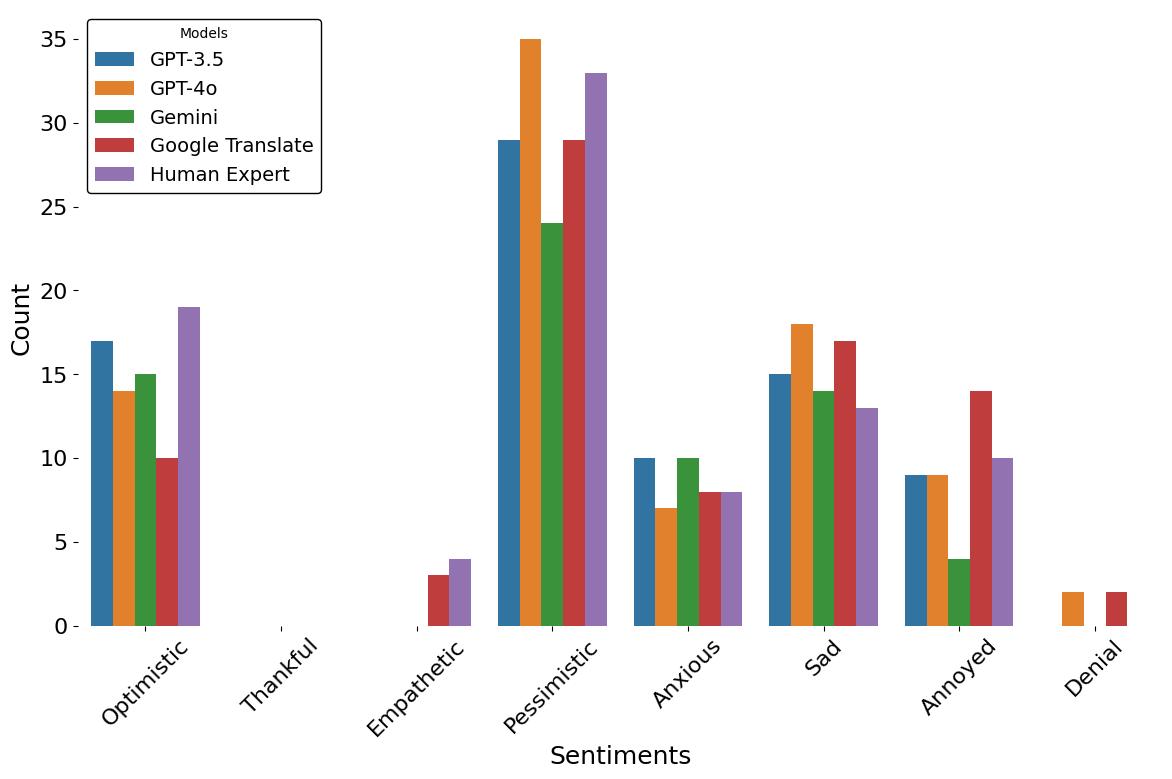}
        \caption{Sentiment Analysis from Google Translate, LLMs and Human Expert  translations (Telugu-English).}
        \label{fig:Telugusentiments}
\end{figure}

Figure \ref{fig:Telugusentiments} presents the sentiments detected for the Telugu-English translations; GPT-4o and Gemini had the highest 'Pessimistic' sentiment counts (35 and 34), while the Human Expert has a similar count (33). We notice that the 'Optimistic' sentiment is highest in the Human Expert’s translation (19), followed by GPT-3.5 (17). Furthermore, the 'Anxious' and 'Sad' counts are similar across models, with Gemini and GPT-3.5 showing the highest anxiety (10 each). Google Translate has the most 'Empathetic' (3) and 'Denial' (2) sentiments, while all other models had zero in these categories.
Google Translate also had the highest 'Annoyed' sentiment (14), whereas Gemini had the lowest (4). GPT-4o and GPT-3.5 showed nearly identical distributions in the "Annoyed" and "Sad" categories. Overall, GPT-4o and Gemini leaned towards pessimism, while Google Translate introduced more emotional variation. The Human Expert's sentiment distribution remains more balanced, closely aligning with LLMs in most categories.

\begin{figure}[htbp!]
        \centering
        \includegraphics[width=1\linewidth]{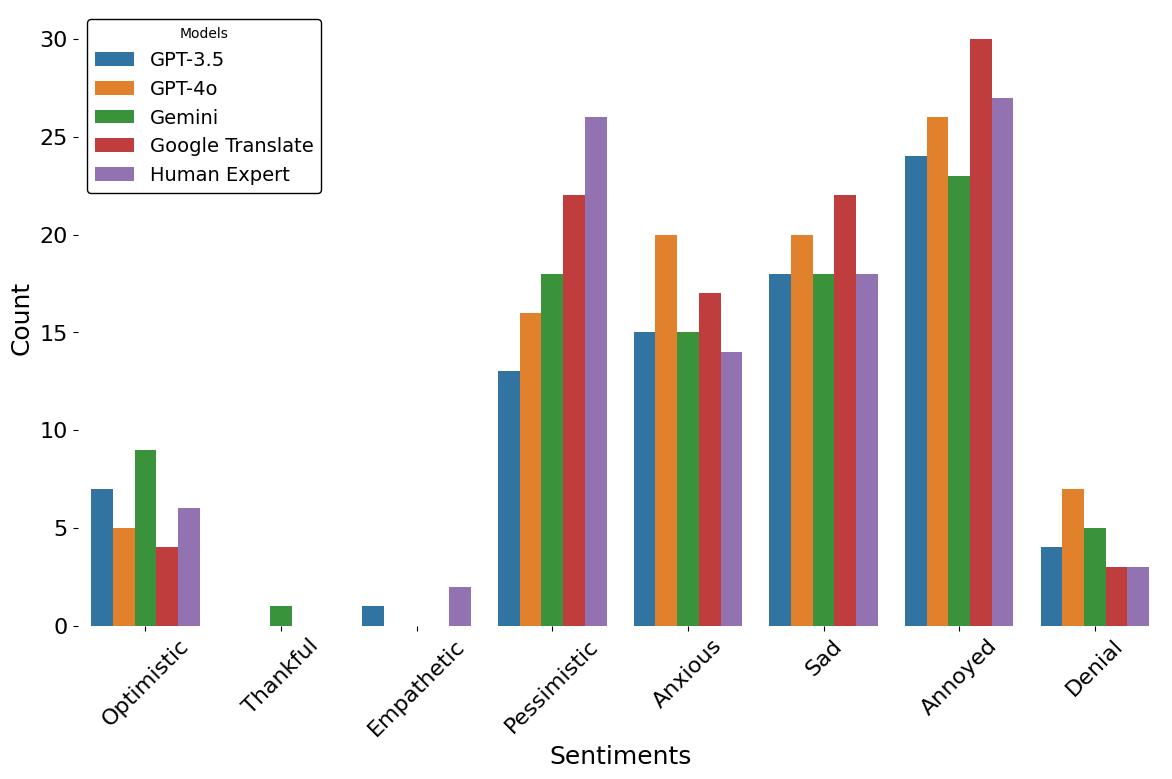}
        \caption{Sentiment Analysis from Google Translate, LLMs and Human Expert  translations (Hindi-English).}
        \label{fig:Hindisentiments}
    \end{figure}

Finally, in the case of the Hindi-English translations (Figure \ref{fig:Hindisentiments}, Gemini produced the highest number of "Optimistic" (9) sentiments, suggesting that its translations may lean toward a slightly more positive tone. In contrast, Google Translate has the lowest count (4), indicating a more neutral or less expressive output. Interestingly, the Human Expert showed the highest number of "Pessimistic" (26) sentiments, implying that human translations might retain more nuance and emotional depth from the original text. Google Translate also featured the highest number of "Sad" (22) and 'Annoyed' (30) sentiments, indicating a potential shift in tone that makes the text sound more negative. Although GPT-4o and the Human Expert had similar 'Annoyed' counts (26 and 27), Gemini remained slightly lower at 23. We find that 'Thankful' is nearly absent from all translations, appearing only once in Gemini’s output. Similarly, we find the 'Empathetic' emotion rarely expressed,  indicating that  LLMs struggle to preserve subtle expressions of empathy. In comparison, we find that the Human Expert producing an 'Empathetic' emotion count of 2. 'Denial' is also relatively low across all translations, with GPT-4o having the highest count (7), possibly due to variations in phrasing and interpretation.

\begin{figure*}[htbp!]
        \centering
        \includegraphics[width=0.8\linewidth]{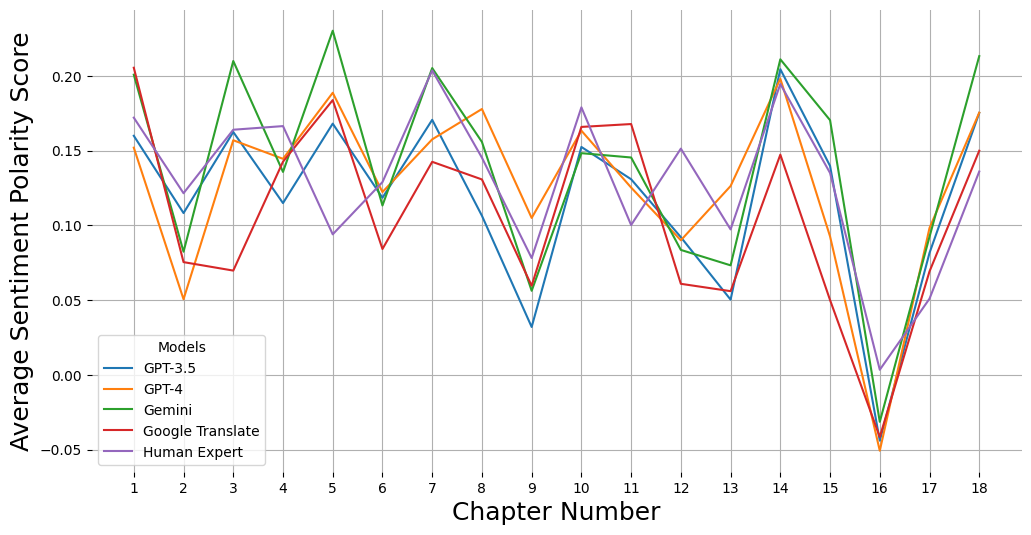}
        \caption{Variation of Sentiment Polarity Scores of models across Chapters (Sanskrit-English)}
        \label{fig:polaritysanskrit}
    \end{figure*}

We next move into sentiment polarity analysis, which provides further details for the entire document.
Figure \ref{fig:polaritysanskrit} presents the sentiment polarity analysis for the  Sanskrit-English translations. The average sentiment polarity scores across 18 chapters show that the models align well with the human expert, exhibiting only minor deviations overall. 

\begin{table*}[h]
    \centering
    \begin{tabular}{l|cc|cc|cc}
        \toprule
        \textbf{Model} & \multicolumn{2}{c|}{\textbf{Sanskrit}} & \multicolumn{2}{c|}{\textbf{Telugu}} & \multicolumn{2}{c}{\textbf{Hindi}} \\
        & \textbf{Mean} & \textbf{Standard Deviation} & \textbf{Mean} & \textbf{Standard Deviation} & \textbf{Mean} & \textbf{Standard Deviation} \\
        \midrule
        GPT-3.5 & 0.124 & 0.283 & 0.083 & 0.308 & -0.012 & 0.178 \\
        GPT-4o & 0.129 & 0.299 & 0.074 & 0.376 & -0.012 & 0.182 \\
        Gemini & 0.143 & 0.299 & -0.008 & 0.263 & -0.028 & 0.196 \\
        Google Translate & 0.114 & 0.296 & -0.001 & 0.358 & -0.04 & 0.203 \\
        Human Expert & 0.131 & 0.299 & 0.097 & 0.283 & -0.087 & 0.226 \\
        \bottomrule
    \end{tabular}
    \caption{\textcolor{black}{Mean and Standard Deviation of TextBlob Sentiment Polarity Score for Different Models.}}
    \label{tab:polarity_table}
\end{table*}

\begin{figure*}[htbp!]
        \centering
        \includegraphics[width=0.8\linewidth]{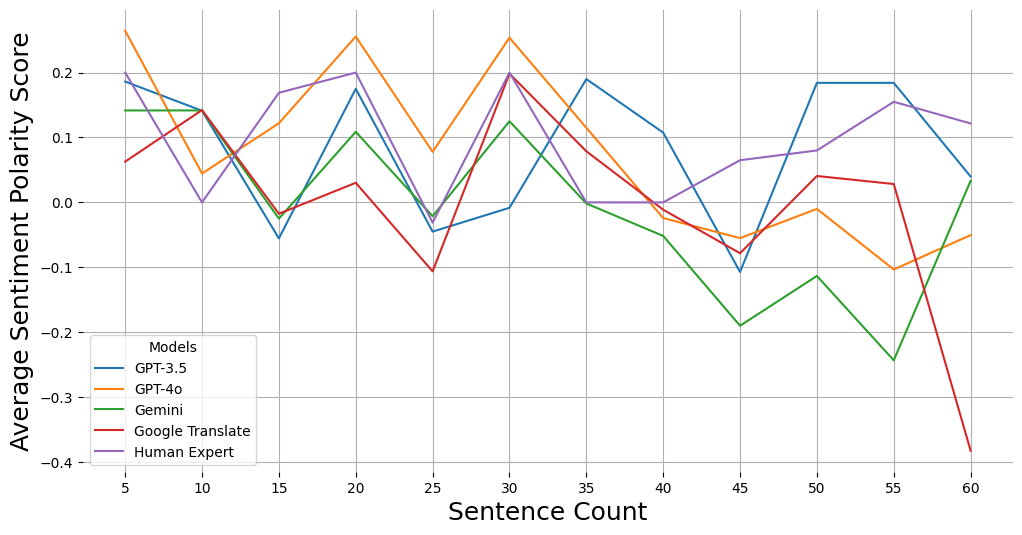}
        \caption{Variation of mean sentiment polarity scores of models across sentences of the Maha P(Telugu-English translation).}
        \label{fig:polaritytelugu}
    \end{figure*}

  Figure \ref{fig:polaritytelugu} presents the Telugu-English translations; overall GPT-4o demonstrates the highest alignment with human expert translations, maintaining relatively consistent polarity scores across sentence counts but deviating towards the end (plus 50 sentences). Gemini follows closely, though it exhibits sharper fluctuations, particularly in the mid-range sentences. GPT-3.5 and Google Translate show significant deviations from the human expert, with Google Translate exhibiting the most erratic behaviour, especially towards the end of the sentence range (plus 50 sentences).
We review the sentences towards the end, where we observe a higher deviation since more figurative elements have been used in the sentences, which might have led to inaccurate translation, and a translation that lacked the sentiment of the actual sentence. 
For example, Sentence 47:  \textit{“On the blade of the dagger, it is like a drop of perspiration” this was the translation by the human expert, and on the other hand, “It looks like a drop of blood on the sword handle”.} We find that the model has translated the sentence when it suggested the tension, fragility by the image of the drop of perspiration.

\begin{figure*}[htbp!]
        \centering
        \includegraphics[width=0.8\linewidth]{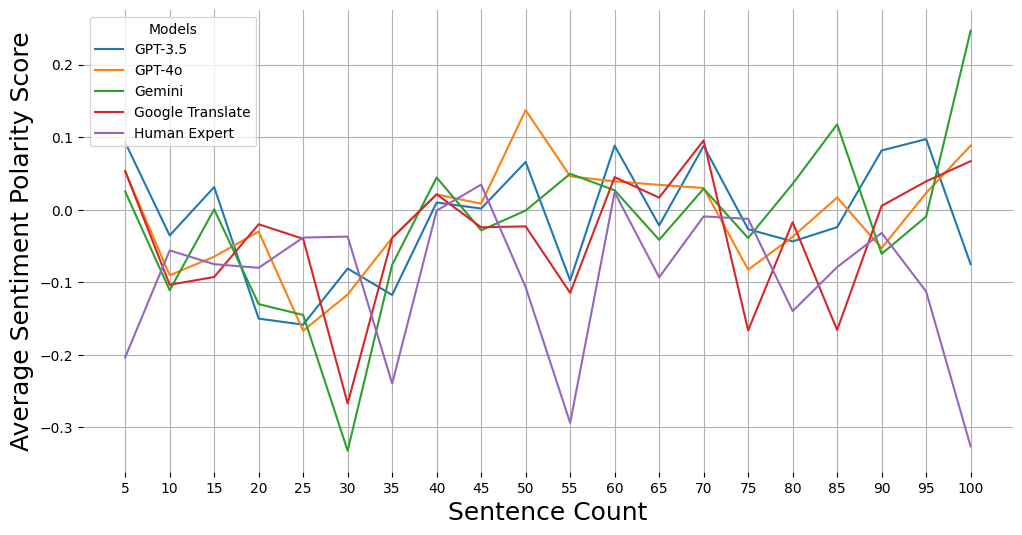}
        \caption{Variation of sentiment polarity scores of models across sentences (Hindi-English translation). }
        \label{fig:polarityhindi}
    \end{figure*}

Similarly, for the Hindi-English translation, Figure \ref{fig:polarityhindi} shows that the models often failed to capture the deep negativity present in human expert translations, as seen for sentences 35, 55, and 65. Although human versions reflect strong emotional shifts, especially in sorrowful or intense passages, the language translation models tend to soften and neutralise these emotions. 


\subsection{Semantic Analysis}\label{lca}

We use the MPNet-Base model\cite{shukla_an}, providing semantic analysis for comparing different translations by  LLMs to a human expert. We evaluate the translations by  GPT-4o, GPT-3.5, Google Translate and  Gemini using the cosine similarity scores (from 0 to 1), where 1 implies a perfect semantic match. We first provide semantic similarity analysis of the respective language translation models across Sanskrit to English translations as shown in Table \ref{tab:sanskritsimilarity}. We observe that GPT-4o achieves the highest average similarity of 0.686, followed by GPT-3.5 at 0.665 and Gemini at 0.660. Although GPT-3.5 attains the highest peak similarity of 0.979, GPT-4o maintains a more consistent performance. Additionally, Google Translate lags with an average similarity of 0.603. Table \ref{tab:chapterwisesanskrit} presents further details with chapter-wise similarity scores, where GPT-4o provides the best scores. Table \ref{tab:sanskritsimilarity} presents further details where the Sanskrit-English translation has been compared.

In the case of Telugu-English translation (Table \ref{tab:telugusimilarity}, Gemini leads with an average similarity of 0.726, outperforming all other models. GPT-4o follows with 0.704, while GPT-3.5 records a lower score of 0.583, showing lower similarity to the human expert translations. Google Translate performs relatively better in Telugu compared to Sanskrit, securing an average similarity of 0.656, but still trails behind the leading models.

Finally, in the case of  Hindi-English translations (Table \ref{tab:hindix})
GPT-4o achieves the highest similarity to human translations, with an average cosine similarity of 0.754 and a peak of 0.927, indicating strong alignment with human understanding. Gemini followed with 0.739, performing slightly better than GPT-3.5 (0.711), but lower than GPT-4o. Google Translate has the lowest average similarity (0.684), suggesting that its translations are less accurate overall. However, its highest similarity score (0.916) was close to that of other models, meaning it could sometimes produce accurate results. The lowest similarity scores highlight consistency differences. GPT-4o has the highest minimum similarity (0.406), while GPT-3.5 dropped to 0.317, showing that its translations are sometimes less reliable.

\begin{table}[h]
    \centering
    \small
    \renewcommand{\arraystretch}{1.2}  
    \setlength{\tabcolsep}{4pt}        
    \begin{tabular}{|p{1.2cm}|p{1.625cm}|p{1.625cm}|p{1.625cm}|p{1.625cm}|}  
        \hline
        \textbf{Chapters} & \textbf{GPT-4o-HE} & \textbf{GPT-3.5-HE} & \textbf{Gemini-HE} & \textbf{GT-HE} \\
        \hline
        1  & 0.667(0.136)  & 0.651(0.145)  & 0.640(0.147)  & 0.574(0.139)  \\
        2  & 0.680(0.150)  & 0.650(0.139)  & 0.622(0.120)  & 0.576(0.150) \\
        3 & 0.736(0.102) & 0.698(0.115) & 0.688(0.109) & 0.583(0.148) \\
        4 & 0.704(0.120) & 0.662(0.127) & 0.678(0.123) & 0.580(0.134) \\
        5 & 0.680(0.116) & 0.639(0.114) & 0.673(0.138) & 0.573(0.118) \\
        6 & 0.640(0.120) & 0.629(0.137) & 0.610(0.112) & 0.556(0.103) \\
        7 & 0.744(0.098) & 0.746(0.112) & 0.685(0.102) & 0.620(0.136) \\
        8 & 0.686(0.088) & 0.649(0.104) & 0.658(0.103) & 0.605(0.133) \\
        9 & 0.749(0.084) & 0.693(0.121) & 0.737(0.088) & 0.614(0.099) \\
        10 & 0.756(0.127) & 0.753(0.131) & 0.677(0.124) & 0.675(0.135) \\
        11 & 0.744(0.101) & 0.711(0.124) & 0.691(0.113) & 0.642(0.112) \\
        12 & 0.705(0.100) & 0.678(0.109) & 0.664(0.118) & 0.610(0.094) \\
        13 & 0.695(0.113) & 0.654(0.129) & 0.673(0.127) & 0.614(0.107) \\
        14 & 0.655(0.134) & 0.674(0.129) & 0.590(0.122) & 0.616(0.103) \\
        15 & 0.684(0.089) & 0.648(0.106) & 0.643(0.093) & 0.573(0.078) \\
        16 & 0.715(0.104) & 0.644(0.129) & 0.663(0.115) & 0.602(0.126) \\
        17 & 0.726(0.080) & 0.676(0.156) & 0.658(0.129) & 0.548(0.153) \\
        18 & 0.688(0.123) & 0.664(0.135) & 0.635(0.122) & 0.590(0.142) \\
        \hline
        \textbf{Average} & \textbf{0.686(0.110)}  & \textbf{0.665(0.126)} & \textbf{0.660(0.117)} & \textbf{0.603(0.123)} \\
        \hline
    \end{tabular}
    \caption{Chapter-wise cosine similarity between language models and Human Expert for Sanskrit-English translation.}
    \label{tab:chapterwisesanskrit}
\end{table}

\begin{table}[htbp!]
    \centering
    \small
    \renewcommand{\arraystretch}{1.5} 
    
    \begin{tabular}{lccc}
        \toprule
        & Average & Highest & Lowest \\
        \midrule
        GPT-4o - Human Expert & 0.686 & 0.943 & 0.106 \\
        GPT-3.5 - Human Expert & 0.665 & 0.979 & 0.201 \\
        Gemini - Human Expert & 0.660 & 0.921 & 0.164 \\
        GT - Human Expert & 0.603 & 0.965 & 0.150 \\
        \bottomrule
    \end{tabular}
    \caption{Semantic analysis showing cosine similarity scores (Sanskrit-English translation).}
    \label{tab:sanskritsimilarity}
\end{table}

\begin{table}[htbp!]
    \centering
     \renewcommand{\arraystretch}{1.5} 
    
    \begin{tabular}{lccc}
        \toprule
        & Average & Highest & Lowest \\
        \midrule
        GPT-4o - Human Expert & 0.704 & 0.907 & 0.354 \\
        GPT-3.5 - Human Expert & 0.583 & 0.908 & 0.321 \\
        Gemini - Human Expert & 0.726 & 0.934 & 0.216 \\
        GT - Human Expert & 0.656 & 0.873 & 0.194 \\
        \bottomrule
    \end{tabular}
    \caption{Semantic analysis showing cosine similarity scores (Telugu-English translation).}
    \label{tab:telugusimilarity}
\end{table}

\begin{table}[htbp!]
    \centering
    \renewcommand{\arraystretch}{1.5} 

    \begin{tabular}{lccc}
        \toprule
        & Average & Highest & Lowest \\
        \midrule
        GPT-4o - Human Expert & 0.754 & 0.927 & 0.406 \\
        GPT-3.5 - Human Expert & 0.711 & 0.912 & 0.317 \\
        Gemini - Human Expert & 0.739 & 0.914 & 0.364 \\
        GT - Human Expert & 0.684 & 0.916 & 0.352 \\
        \bottomrule
    \end{tabular}
    \caption{Semantic analysis showing cosine similarity scores (Hindi-English translation).}
    \label{tab:hindix}
\end{table}

We further evaluate the accuracy of Hindi and Telugu translations,  with the mean and standard deviation of the similarity scores. We group the sentences into bins of  10 sentences to calculate the mean similarity score and standard deviation.

\begin{figure*}[htbp!]
        \centering
        \includegraphics[width=0.8\linewidth]{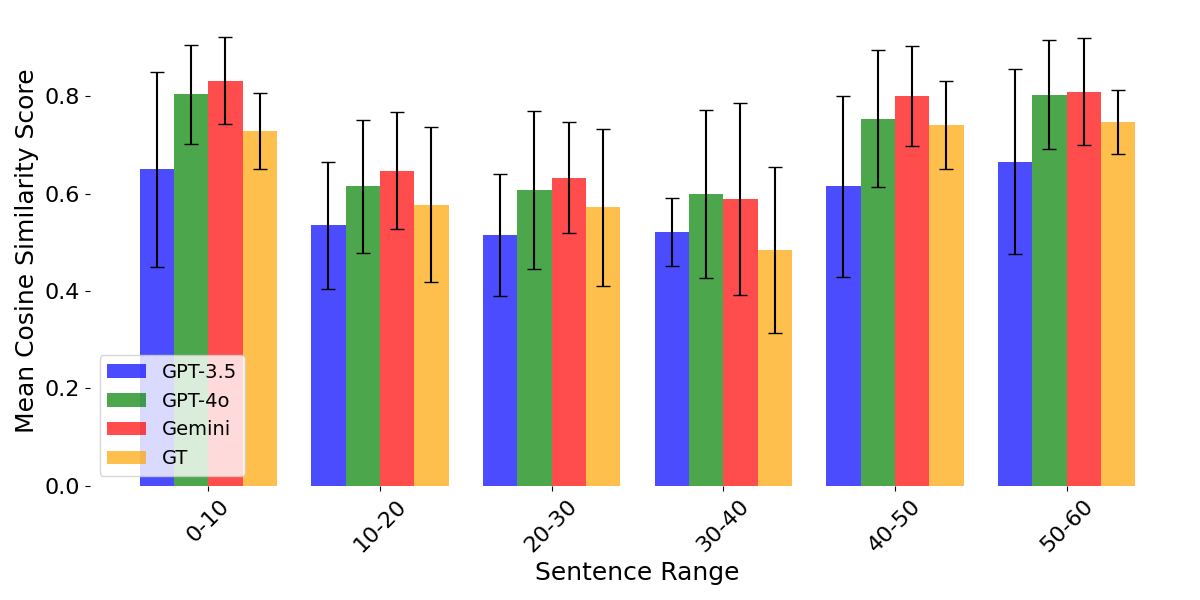}
        \caption{Mean Similarity Score vs Sentences (Telugu)}
        \label{fig:errorbartelugu}
    \end{figure*}

In the case of Telugu-English translation (
Figure \ref{fig:errorbartelugu}), we find that the  the similarity scores declined significantly across all models for sentences in the 10–40 range. These sentences contained extensive use of personification, imagery, symbolism, and abstract elements, which posed challenges for translation. Examples of such sentences include: “Roar forth, oh ocean of flames!” and “The mountain moves like a massive chariot!” The complexity of these expressions contributed to the lower similarity scores. Despite this, GPT-4o and Gemini remained the most reliable models, whereas GPT-3.5 exhibited significant inconsistencies.

\begin{figure*}[htbp!]
        \centering
        \includegraphics[width=0.8\linewidth]{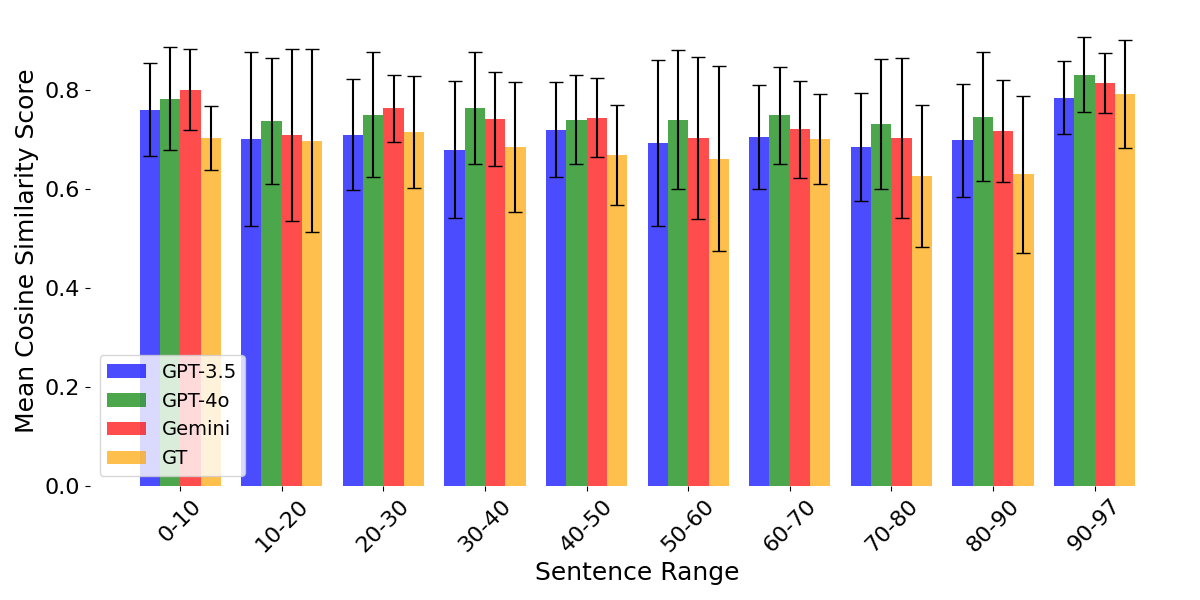}
        \caption{Mean Similarity Score vs Sentences (Hindi)}
        \label{fig:errorbarhindi}
    \end{figure*}

In the case of Hindi-English translation (
Figure \ref{fig:errorbarhindi}), we observe that sentences in the 1–10 and 90–97 ranges exhibit higher similarity to human translations. A closer analysis of the translated texts revealed that these sentences primarily consist of clear and direct narration with minimal use of figurative language. Furthermore, the sentences in the 90–97 range demonstrate the least variability, as reflected in Figure   \ref{fig:errorbarhindi}) for similar reasons. Among the models, GPT-4o and Gemini consistently outperform others when compared to Human Expert translation.

The semantic similarity analysis of language translation models across Sanskrit, Telugu, and Hindi highlights that GPT-4o is the most consistent and reliable model, performing well in all languages, particularly in Sanskrit and Hindi. Gemini excels in Telugu but struggles with Sanskrit and Hindi, suggesting it’s more suited for certain languages. GPT-3.5 shows high peak similarity in Sanskrit but lacks consistency. Google Translate consistently underperforms, especially in Sanskrit and Hindi, indicating limitations in handling these languages’ complexities. Overall, GPT-4o is the most dependable, while Gemini shines in Telugu and Google Translate lags behind.

\subsection{Qualitative analysis of  semantically similar verses }

 We qualitatively evaluate how well the translation has been done for the semantically most similar verses and least similar verses to give us an insight into how well the models are performing and whether they correctly transmit the original meaning of the text.

As shown in Table \ref{tab:mostleastsanskrit}, in general, we find that the respective language models show high accuracy in translating structured, fact-based Sanskrit (Bhagavad Gita) statements, while struggling with abstract, poetic, and philosophical content. We find that the sentences with direct relationships and clear syntactic patterns, such as enumerations (Chapter 10, Verse 31): \textit{"Of the purifiers, I am the wind; of the wielders of weapons, I am Rama"}), achieve the highest similarity scores (close to 0.90). In contrast, abstract and philosophical sentences discussing concepts such as the imperishability of the soul or renunciation score significantly lower (around 0.10–0.40) due to their metaphorical language and cultural references.  The emotional and metaphoric language expressions bring further challenges to language models with verses such as  Arjuna’s distress in battle  (Chapter 1, verse 28): \textit{"my limbs fail and my mouth is drying up,}" receive low similarity scores (0.36 for GPT-4o, 0.37 for GPT-3.5), as models struggle to convey the intended sentiment. Similarly, we find that honorifics and formal expressions are often inconsistently translated, with phrases such as \textit{"O best of the Indians"} being altered, affecting both tone and cultural accuracy. These inconsistencies highlight the limitations of current language models in handling context-dependent translations. Furthermore, GPT-4o and GPT-3.5 perform well on structured statements but struggle with literary and philosophical Sanskrit texts (Chapter 2, verse 30).

In the case of Telugu-English translation (Table \ref{tab:mostleasttelugu}, sentences with a clear structure, like those describing a lamp turning into a star, scored the highest across all models. GPT-4o, GPT-3.5, and Gemini all had similarity scores above 0.9, indicating they effectively captured the sequence of events. The minor differences in phrasing indicate variations in sentence construction rather than a loss of meaning, as seen in verse 48. The Human Expert translation states: \textit{"The star sang its inviting song. The lamp went out. It transformed into a star."} Meanwhile, the Google translation reads: \textit{"The celestial star sang an invocation. The lamp went out. Turned into a star."} Despite differences in wording, the overall meaning remains unchanged. The descriptive and philosophical sentences, such as those discussing the vastness of the sky or the feeling of disappearing into the cosmos, as described in sentences 33 and 14, respectively. They showed moderate similarity (0.354–0.520), which relies on figurative language, which the models often interpret differently, leading to minor shifts in meaning. GPT-4o was consistent in maintaining the original tone, though some rewording was observed.

In the case of Hindi-English translation (Tamas) shown in Table \ref{tab:mostleasthindi}, structured and action-oriented sentences show the highest similarity, with GPT-4o scoring 0.927. However, figurative and descriptive elements resulted in lower scores. For example (Sentence 12),  in \textit{"thorny bristles on its back,"} the models struggled to retain details, leading to scores of 0.520 (GPT-4o), 0.317 (GPT-3.5), and 0.364 (Gemini). Similarly, metaphorical expressions such as \textit{"he cursed the evil moment"} had lower scores: 0.536 (GPT-4o), 0.535 (GPT-3.5), and 0.498 (Gemini). The sensory descriptions, such as the pig’s movement, showed moderate similarity (0.919 for GPT-4o), suggesting that while models captured the overall meaning, they missed finer details. Emotionally charged phrases, such as \textit{"the dagger in his hand looked odd and irrelevant,"} showed greater divergence, with scores of 0.785 (GPT-4o), 0.655 (GPT-3.5), and 0.700 (Gemini).

\begin{table*}[h!]
    \centering
    \renewcommand{\arraystretch}{1.2}
    {\fontsize{7}{7}\selectfont
    \begin{tabular}{|p{0.8cm}|p{0.8cm}|p{1.5cm}|p{1.5cm}|p{1.5cm}|p{1.5cm}|p{1.5cm}|p{0.8cm}|p{0.8cm}|p{0.8cm}|p{0.8cm}|}
        \hline
        \textbf{Chapter} & \textbf{Verse} & \textbf{HumanExpert} & \textbf{GPT-4o} & \textbf{GPT-3.5} & \textbf{Gemini} & \textbf{GT} & \textbf{GPT-4o-HE} & \textbf{GPT-3.5-HE} & \textbf{Gemini-HE} & \textbf{GT-HE} \\
        \hline
          10 & 31 & I am the Wind among purifiers, the King Rama among warriors; I am the Crocodile among the fishes, and I am the Ganges among rivers.   & Of the purifiers, I am the wind; of the wielders of weapons, I am Rama; of the fishes, I am the makara; of the rivers, I am the Ganges. & I am the wind among the purifiers, I am Rama among the warriors, I am the crocodile among the fish, and I am the Ganges among the rivers.
          & Among purifiers, I am the wind (Pavan); among those who wield weapons, I am Rama; among fish, I am the Makara; and among rivers, I am the Ganga. 
          & Among the winds among the winds I am Rama among the bearers of weapons I am the crocodile among the fish I am the source of the streams I am the Jahnavi. & 0.904 & 0.979 & 0.851 & 0.870 \\
        \hline
          7 & 10 & Know, O Arjuna, that I am the eternal Seed of being; I am the Intelligence of the intelligent, the Splendour of the resplendent. 
          & Know me, O Partha, as the eternal seed of all beings. I am the intelligence of the intelligent, and the brilliance of the brilliant. 
          & Know Me to be the eternal seed of all beings, O Arjuna. I am the intelligence of the intelligent, and the splendor of the splendid. & Know Me, O Partha, as the eternal seed of all beings. I am the intellect of the intelligent, the brilliance of the brilliant.
         & 
         O son of Pṛthā, know Me to be the eternal seed of all living entities. .

         & 0.804 & 0.965 & 0.722 & 0.707 \\
        \hline
         10 & 4 & Intelligence, wisdom, non-illusion, forgiveness, truth, self-control, calmness, pleasure, pain, birth, death, fear and fearlessness; 
         & Intelligence, knowledge, non-delusion, forgiveness, truth, self-restraint, calmness, happiness, sorrow, existence, non-existence, fear, and also fearlessness.
         & Intelligence, knowledge, non-delusion, forgiveness, truth, control over the senses, control over the mind, pleasure, pain, birth, death, fear, and fearlessness.
         & Intellect, knowledge, freedom from delusion, forgiveness, truthfulness, self-control, peace, pleasure, pain, existence, non-existence, fear, and fearlessness - these are My different manifestations.
         & Intelligence, knowledge, non-delusion, forgiveness, truthfulness, restraint, happiness, pain, birth, fear and fearlessness.
         & 0.943 & 0.940 & 0.679 & 0.965 \\
        \hline
         2 & 30 & Be not anxious about these armies. The Spirit in man is imperishable.  & The soul is eternally indestructible in the body of all, O Bharata (Arjuna). Therefore, you should not grieve for any creature. & The embodied soul is eternal and indestructible, O Arjuna; therefore, you should not grieve for any creature. & The Blessed Lord (Krishna) said:
         "O Bharata, the Self (Atman) that dwells in the body is eternal and indestructible. Therefore, you should not grieve for any being." 
         & O descendant of Bharata, the embodied soul is always invincible to the body. Therefore, O descendant of Bharata, you should not grieve for all living beings.  & 0.106 & 0.234 & 0.164 & 0.211 \\
        \hline
        18 & 4 & O best of Indians! Listen to my judgment as regards this problem. It has a threefold aspect. & Hear from me the definite conclusion about renunciation, O best of the Bharatas. Renunciation, O tiger among men, is indeed declared to be of three kinds. & Listen to my determination regarding renunciation, O best of the Bharatas. Renunciation is declared to be of three kinds. & O best of the Bharatas, O tiger among men, hear from me about the settled conclusion on relinquishment. Relinquishment is of three kinds, as I will now describe. &
        O best of men, O best of men, I have now described to you in three ways, O best of men. & 0.346
        & 0.427 & 0.402 & 0.150 \\
        \hline
        1 & 28 & My limbs fail me and my throat is parched, my body trembles and my hair stands on end. 
          & Arjuna said: Seeing these, my own kinsmen, O Krishna, arrayed and eager for battle, my limbs give way and my mouth is drying up. & Arjuna said: O Krishna, seeing my own kinsmen standing arrayed, eager for battle, my limbs fail and my mouth is parched.
         & Arjuna said: "O Krishna, seeing this kin of mine, eager to fight, standing before me, my limbs tremble and my mouth dries up." 
         & Arjuna said: O Kṛṣṇa, when I saw this relative of this people, my limbs were traced to fight. 28. & 0.364 & 0.372 & 0.285 & 0.187 \\
         \hline
    \end{tabular}
    }
    \caption{Selected most and least semantically similar verses from the Bhagavad Gita (Sanskrit-English) translation.}

    \label{tab:mostleastsanskrit}
\end{table*}

\begin{table*}[htbp!]
    \centering
    \renewcommand{\arraystretch}{1.2}
    {\fontsize{7}{7}\selectfont
    \begin{tabular}{|p{0.6cm}|p{1.9cm}|p{1.9cm}|p{1.9cm}|p{1.9cm}|p{1.9cm}|p{1cm}|p{1cm}|p{1cm}|p{1cm}|}
        \hline
r        \textbf{Verse} & \textbf{HumanExpert} & \textbf{GPT-4o} & \textbf{GPT-3.5} & \textbf{Gemini} & \textbf{GT} & \textbf{GPT-4o-HE} & \textbf{GPT-3.5-HE} & \textbf{Gemini-HE} & \textbf{GT-HE} \\
        \hline
          48 & Suddenly, that lamp saw a star in the sky. It cried out like a she-demon. (Cries that would pierce your ears and mine) The star showered its affectionate glances. The poor lamp grew weary. The star sang its inviting song. The lamp went out. It transformed into a star.  & 
          Suddenly, the lamp looked at the star in the sky. The demon screamed. (Screams that neither you nor I can hear) The star in the sky extended a loving gaze. Poor thing, the lamp was exhausted. The star in the sky sang a welcoming song. The lamp went out. It turned into a star.
          & Unexpectedly, that lamp. Saw the stars in the sky. It wore a demon's attire. The stars shone brightly with respect. Alas, it went out, the lamp. The stars beckoned it. The lamp flickered. It turned into a star.
          & Suddenly that lamp looked at the star in the sky. It screamed like a demon. (Screams that cannot reach our ears) The star cast admiring glances. The lamp was tired, poor thing. The star invited. The lamp went out. It turned into a star. 
          & Suddenly the lamp saw the celestial star. The rakasi cried out. (Crying that did not infect the ears of you and me) The celestial star extended a loving glance. Exhausted, poor, the lamp. The celestial star sang an invocation. The lamp went out. Turned into a star.
          & 0.906 & 0.908 & 0.934 & 0.838 \\
        \hline
          1 & Another world, another world, another world has called! March forward, march onward! Let's go, let's go upward! 
          & Another world, another world, another world called! Let's go forward, let's go ahead! Let's go, let's go upwards! 
          & Another world, another world, another world has emerged! Step forward, step backward! Let's go, let's go up!
          & Another world, another world, another world is calling! Come forward, push forward! Let's go, let's go upwards!
          & 
           Another world, another world, another world called! Come on, come on! Let's go, let's go!

         & 0.906 & 0.841 & 0.924 & 0.796 \\
        \hline
         56 & A puppy, a spark, a bubble - do not look down upon anything! Everything is poetic! A bread crumb, a drop of sweat, a small fish - they all gaze at you! They ask you to fathom their depths! A doorknob, a bead necklace, a horse's anklet - none of these is unworthy of poetry! Each is a priceless sculpture! One must possess poetic frenzy! But one must also have the direction of aesthetic taste! Failing which, not a hint of beauty will be found! If you have eyes, see and write! If you have speech, speak and write! The world is a lotus array! Poetry is an unquenchable thirst!
         & Puppy, matchstick, soap bubble - Do not look down on anything! Everything is filled with poetry! Bread slice, banana peel, wooden plank - They keep looking at you! They ask you to understand their depth! Door latch, offering plate, horse bridle - Nothing is unworthy of poetry! Yes, indeed, sculpture is priceless! Let the poetic fervor remain! Let your taste guide you! The splendor is said to be unobtainable! If you have eyes, see, if you have words, write! The world is a labyrinth! Poetry is an unquenchable thirst!
         & Little puppies, tiny embers, burnt matches, Don't see these as lowly!  All that's poetic,  Even in crumbs, shards, pieces of grass,  They exist while looking at you!  We can't ignore their beauty! The glow of a dewdrop, a dew's tear, a pearl's reflection, Isn't it worthy of poetry? A craft without an artist!  The spirit of poetry itself!  A direction without taste!  Beauty without being found!  Look at the stones,  Speak if you can!  The world is a lotus formation! Poetry is an unquenchable thirst!
         & Don't look down on anything, puppy, firefly, soap bubble! Everything is poetic! A piece of bread, a banana peel, a piece of wood - they all look at you! They ask you to find their depth! Door knob, offering plate, horse's eyes - nothing is unworthy of poetry! Yes, yes, sculpture is worthy! There should be poetic inspiration! But you are the direction of rasa! The beauty of it is not available! If you have eyes, look, if you have words, write! The world is a lotus maze! Poetry is an insatiable thirst!
         &  A puppy, a firecracker, a bar of soap- Don't look at anything bad! Everything is poetic! A piece of bread, a banana, a wooden stick- Looking at you! Let's find their depth! A doorknob, a hearth plate, a horse bone- Doesn't deserve a poet! Naunu Shilpamanargham! Undaloi Kavitavesham! Kanvoi Rasa Nirdesh! Dorakadatoi Shobhalesham! See if you can see, Write if you want! World-class Padmavyuham! Poetry-like insatiable thirst!
         & 0.889 & 0.845 & 0.910 & 0.858 \\
        \hline
         33 & In the horizon of the sky, the meteor is left alone like an abandoned wayfarer!
         & The moon looks like a lone camel with broken legs in the sky's vastness!
         & In the sky's playground, clouds play hide and seek like children!
         & The jackal looks like a lonely camel with broken legs in the desert of the sky!
         & Zabilli is like a lonely camel with its legs cut off in the sky! & 0.354 & 0.417 & 0.216 & 0.194 \\
        \hline
        19 & If I fill myself in the cosmic womb, those moments when my whirlpools make the world drift would arrive!
        & I am filled with joy throughout the world, my whispers spread across the entire universe, those moments are captivating!
        & When I close my eyes, I disappear, my shadow spreads like a mirage, the world seems like waves!
        & I fill the whole earth, and the peaks of my caves bring about the moment when the whole world will be showered like rain! &
        My neck is filled with dhatri, the sikharas of my vulva are like that moment when the whole world will be showered! & 0.395
        & 0.394 & 0.520 & 0.302 \\
        \hline
         14 & Didn't I pant like a fanning bellows when summer blazed?
          & Didn't you curl up like a bat when the summer heat was scorching? & When the time ends, the sky will not cry like a dove!
         & Didn't I burn like a sparrow when the summer was hot? 
         & Don't you fly like a bat when the sun burns! & 0.564 & 0.321 & 0.757 & 0.474 \\
         \hline
    \end{tabular}
    }
    \caption{Selected most and least semantic similar verses from the Maha Prashatham (Telugu-English translation). 
}
    \label{tab:mostleasttelugu}
\end{table*}

\begin{table*}[htbp!]
    \centering
    \renewcommand{\arraystretch}{1.2}
    {\fontsize{7}{7}\selectfont
    \begin{tabular}{|p{0.6cm}|p{1.9cm}|p{1.9cm}|p{1.9cm}|p{1.9cm}|p{1.9cm}|p{1cm}|p{1cm}|p{1cm}|p{1cm}|}
        \hline
        \textbf{Verse} & \textbf{HumanExpert} & \textbf{GPT-4o} & \textbf{GPT-3.5} & \textbf{Gemini} & \textbf{GT} & \textbf{GPT-4o-HE} & \textbf{GPT-3.5-HE} & \textbf{Gemini-HE} & \textbf{GT-HE} \\
        \hline
          10 & He had struck the
          pig several times on its belly and shoulders, but on pulling out the knife,
          only a few drops of blood would ooze out and trickle down to the floor.
          There was no trace of any wound or stab. A thin, red line of a scratch or a tiny blot were the only visible signs of his efforts. 
          & 
          He had stabbed the pig's belly and shoulders countless times. When he pulled out the knife, a few drops of blood would fall on the floor, but at the site of the wound, only a small line or a tiny spot remained, which wasn't even visible on the pig's skin.
          & 
          He had stabbed the pig's belly and shoulders several times. When he pulled out the knife, a few drops of blood fell on the floor, but instead of a wound, there was a small leak or a small stain filling up, which was not visible on the pig's skin.
          & 
          He had stabbed the pig countless times in its belly and shoulders.  When he removed the knife, a few drops of blood would fall on the floor, but the wound would only leave a tiny leak or small stain barely visible in the pig's skin.
          & 
          How many times he had stabbed the stomach of the pig and on the shoulders.When the stabbing was removed, a few drops fell on the floor of the blood, but instead of the wound, there would be a small leak or small spot which did not even appear in the pig's skin.
          & 0.927
          & 0.900 
          & 0.914 
          & 0.800 \\
        \hline
          31 & Nathu could not refuse either. How could he? He dealt with Murad Ali
          almost every day. Whenever a horse or a cow or a buffalo died anywhere in
          the town, Murad Ali would get it for him to skin. It meant giving an eightanna piece or a rupee to Murad Ali but Nathu would get the hide.
          & 
          Nathu couldn't refuse Murad Ali, as he needed him daily. Whenever a horse died in the city, or a cow or buffalo died, Murad Ali would arrange to get the hide, and Nathu had to give Murad Ali half a rupee or a rupee, but he would get the hide.
          & 
          Murad Ali used to work for Nattu every day, how could Nattu refuse. Whenever a horse died in the city, or a cow or buffalo died, Murad Ali would arrange for the skin to be delivered, Nattu also had to pay eight annas to Murad Ali, but the skin would be obtained.
          & 
          Murad Ali was needed every day, how could Natthu refuse him? Whenever a horse, cow, or buffalo died in the city, Murad Ali would arrange for the hide to be obtained.  Murad Ali had to be paid as well, but the hide would be obtained.
          &
          Murad Ali used to work daily, how would Nathu refuse.Whenever a horse would die in the city, a cow or buffalo died, Murad Ali used to get the skins, the eighty rupee Murad Ali would also have to give it, but the skins would have been found.

          & 0.921 
          & 0.722 
          & 0.780 
          & 0.881 \\
        \hline
         94 & The pig had actually moved. It walked ponderously towards Nathu. It had hardly
         taken a couple of steps when it staggered and emitted a strange sound.
         & 
         The pig really moved. It truly began to advance towards Natthu with a sluggish, steady pace. After taking a couple of steps, swaying side to side, a strange sound emerged from the pig's mouth.
         & 
         The pig was really shaking. It was truly moving forward with a heavy, steady pace, heading towards the hole. After taking a couple of steps, swaying left and right, a strange sound came out of the pig's mouth.
         & 
         The boar was truly agitated.  It really began to move forward, towards Natthu, at a heavy, slow pace. After a few steps, swaying left and right, a strange sound emerged from the boar's mouth.
         & 
         The pig was really shaken.He was truly moving towards Nathu, a cumbersome stable speed.Two steps, a strange voice came out of the pig's mouth after walking right and left.
         & 0.919 
         & 0.766 
         & 0.777 
         & 0.897 \\
        \hline
        12 & Why of all the creatures, Nathu thought, had this despicable brute with its
        fat, bulging body, a snout covered with thick white hair like that of a
        rabbit’s, thorny bristles on its back, fallen to his lot to tackle?
         &
         Even when I got to kill, it was such a wretched pig. Ugly, with such a big belly, black hair on its back, and white prickly hair around its snout like that of a porcupine.
         & 
         Even if he got a chance to hit, what an unlucky pig. Ugly, with such a big belly, black hair on the back, white prickly hair around the mouth like a thorn.
         & 
         Even if it was found to be killed, what an unlucky pig.  Ugly, such a large belly, black hair on its back, and the hair around its snout is white and prickly like a porcupine's.
         & 
         Even if you got to kill, what kind of wretched pig.The hair on such a large back is black, the hair around the throat is like a white thorny.
         & 0.520 
         & 0.317 
         & 0.364 
         & 0.403 \\
        \hline
         84 & The dagger in
         his hand looked odd and irrelevant. He felt like running away from it all.
         & 
         Coming outside, the knife in his hand seemed incongruous. He felt like running away from there, not even looking into the room, and just running away.
         & 
         Upon coming outside, he felt the knife in his hand and it seemed inappropriate. His mind wanted to flee from there, without even looking inside the room, and run away.
         &
         Coming outside, the knife he held in his hand began to feel incongruous. He felt like running away from there, without even looking inside the cupboard, and just running away.
         &
         After coming out, he started looking uncomfortable.He felt like running away from there, peeping into the closet and does not even run away.
         & 0.785
         & 0.655
         & 0.700 
         & 0.351 \\
        \hline
         16 &  He cursed the evil moment he had agreed to
          do the job. 
          & 
          Who knows in what ill-fated moment he took on this task.
          & 
          Who knows in what wrong time he took this task upon himself.
          & 
          Who knows at what wrong time he took on this task.
          & 
          Know what he had taken this work on his head.
          & 0.536 
          & 0.535 
          & 0.498 
          & 0.354 \\
         \hline
    \end{tabular}
    }
    \caption{Selected most and least semantically similar verses from Tamas (Hindi-English) translation. 
    }
    \label{tab:mostleasthindi}
\end{table*}

\section{Discussion}\label{evl}

We first summarise the results from the different investigations that include sentiment and semantic analysis for comparing the respective translations of the given texts (Bhagavad Gita, Tamas, and Maha Prasthanan).
The sentiment analysis results show that pessimistic, anxious, and sad tones appear across all models in the Sanskrit-English translation (Figure 2), reflecting the solemn and philosophical nature of the Bhagavad Gita. However, sentiment distribution varied significantly between models (Figures 6 and 7). GPT-3.5 reported the closest sentiment polarity scores when compared to the human expert (Table 9). Considering the Hindi-English translation of Tamas, the Human Expert version had the most pessimistic sentiments, indicating a stronger focus on negative tones (Figure 4). GPT-4o showed the highest Anxious sentiment, suggesting it may amplify emotional intensity. Google Translate had the most Sad and Annoyed sentiments, reflecting a generally negative tone, but also expressed the most empathy when compared to Human expert and other models.  Gemini had the highest Optimistic sentiment, making its translations more positive in comparison. In Telugu translation (Figure 3), sentiment distribution is relatively more consistent across models. However, Google Translate introduced more emotional variation, generating translations that show empathy and denial. GPT-4o and Google Translate leaned toward pessimism, similar to their Hindi translations. This consistency suggests that while LLMs can recognise broad emotional trends, they still exhibit model-specific biases in sentiment interpretation. In general, all models have been close to the Human expert for most of the sentiments, except for empathetic sentiments in Telugu-English (Figure 3) and Hindi-English (Figure 4) translations. These models have captured the Empathetic sentiment better in the case of the Sanskrit-English translation (Figure 2). The difference in performance could be due to the nature of the texts, i.e more empathetic sentiments were expressed in the Bhagavad Gita when compared to Tamas and Maha Prasthanam. Furthermore, we have a larger text data size in the case of the Bhagavad Gita, as all 18 chapters were considered when compared to selected chapters from Tamas and Maha Prasthanam.

The sentiment analysis highlights the importance of context in translation. Many variations suggest that LLMs translate text into English without integrating background knowledge from the source material. Providing contextual prompts could enhance translation quality, ensuring the emotional tone aligns more closely with the original text. Further research could explore whether incorporating historical, cultural, or philosophical context improves sentiment preservation.

The semantic similarity analysis across Sanskrit, Telugu, and Hindi gave varying model performances (Table 10). In Sanskrit, GPT-4o achieved the highest average similarity and maintained consistency, while Google Translate struggled with Sanskrit’s complex syntax and poetic structure. We observed that all the respective models translated structured, fact-based Sanskrit phrases well, but struggled with abstract and philosophical content featuring metaphors. The sentences discussing deep concepts, such as the imperishability of the soul (Table 14), produced lower similarity scores, indicating that metaphorical and figurative language remains a challenge.

In the case of Telugu-English translation (Figure 8), Gemini outperformed other models in overall semantic similarity, followed by GPT-4o. GPT-3.5 scores significantly lower, indicating weaker alignment with human translations. Structured sentences describing sequences of events, such as a lamp turning into a star, achieve high similarity scores, suggesting that LLMs excel at preserving factual meaning. However, figurative expressions about cosmic vastness or abstract emotions show moderate similarity, implying a tendency for models to rephrase rather than directly translate poetic language.

The evaluation of Hindi translations across different models using cosine similarity scores shows that GPT-4o provides the most accurate results (Figure 9), closely matching human translations. It maintains the meaning, sentence structure, and context of the original text better than other models. Gemini also performs well, but sometimes lacks the level of accuracy seen in GPT-4o. GPT-3.5 provides moderate accuracy but tends to simplify certain phrases, which can lead to minor changes in meaning. Google Translate has the lowest accuracy, often struggling with sentence structure and context. The results suggest that as language models improve, they better capture the meaning of the text, with GPT-4o currently being the most reliable for Hindi translation. Overall, GPT-4o emerged as the most reliable model across all three languages, excelling in Sanskrit (Table 10),  and Telugu and Hindi translations (Figures 8 and 9). Gemini performs well in Telugu but struggles with Sanskrit and Hindi, indicating it may be optimised for specific linguistic structures. GPT-3.5 shows strong peak similarity in Sanskrit but lacks overall consistency. Google Translate remains the weakest performer, particularly in Sanskrit and Hindi, highlighting its limitations in handling complex linguistic structures.  However, we note that if we take the standard deviation into consideration (Table 10 and Figures 8 and 9), across all results, no one model is significantly better than another. 

In terms of the limitations, we note that we used translation by one human expert to compare LLMs and Google Translate. The texts such as the Bhagavad Gita have been translated hundreds of times  in the last two millennia, and we note that even the human expert translations have variations when it comes to vocabulary, sentiments and semantics as shown in earlier studies \cite{chandra_2022_semantic}. Furthermore, texts such as the Bhagavad Gita feature the philosophical foundation of Hinduism and are full of allegories and metaphors, which have cultural and religious contexts, making them harder to translate. The composition of the text was more than 2500 years ago and orally transmitted over a millennium before being written down. Therefore, it is not an easy test for translation, both for humans and language models.   Taking these into account, we considered Hindi, which is the largest language (native speaker) in India and Telugu which has over 83 million native speakers (based on the 2011 Indian census) \cite{chandramouli2011census}. We have only considered one text from these languages, and the texts we selected are classical texts (Tamas and Maha Prasthanam), which is a limitation.  Furthermore, we selected only one translation per text since we reviewed three different texts and three LLMs, additional translations can be added in future work.
Further work needs to be done, taking modern texts and more recent works into account, including newspapers and social media, as these are more prominently auto-translated by language models. We also need to consider more human expert translations for effectively evaluating the strength of novel language models for machine translation.  Since  LLMs are black box models, interpretability of how decision was made is a challenging task which depends on model architecture, training strategy and data \cite{csahin2025unlocking}. The way the LLM makes a decision depends mostly on the model architecture and data used in pre-trained Transformer-based models. The training data for the respective LLMs are different, and hence they make slightly different decisions, for example, our results show that GPT-4o maintains emotions better  in translations of philosophical texts such as the Bhagavad Gita.

\textcolor{black}{It is difficult even for human translators to maintain cultural differences and differences in language structure to maintain emotional and semantic retention in translation. Previous study has shown that there is a large difference in the use of vocabulary by human translations of philosophical texts such as the Bhagavad Gita \cite{chandra_2022_semantic}.  However, in the same study, it has also been found that the semantic and emotional similarity were maintained despite the difference in vocabulary.  We note that  different periods in time influence the vocabulary of spoken and written language; therefore, the results depends on what periods (writing style and vocabulary)  contributed to the majority of the training data corpus for the LLM.}

\textcolor{black}{The results from this study about the assessment of LLMs can guide future research in the area of language translation for low-resource languages. The observations from this study, including the strengths and weaknesses of the respective models can be used as a guideline to build better models. Better translation models would require rich and diverse datasets from a wide range of translations and translation experts. Furthermore, a human-in-the loop \cite{wu2022survey} approach would incorporate human translators that can give feedback to LLMs so that it can learn and improve from the mistakes in mistranslations. Furthermore, ablation studies can be done to analyse the individual impact of sentiment versus semantic evaluation to review the similarity and difference of the translations by the different LLMs.
}

Future research could explore improving translations by incorporating contextual background, cultural cues, and domain-specific fine-tuning to enhance both sentiment alignment and semantic accuracy. LLMs offer more advantages over Google Translate and Human Expert since the user can promote LLMs to provide a specific vocabulary style and tone in terms of sentiments. Furthermore, the translated text by human experts can also be transformed by LLMs to get different sets of vocabulary, for example, if the translation of the Bhagavad Gita is from the 1920s, LLMs can be used to give a modern rendering in terms of vocabulary, which would be easier to understand by modern readers. Our framework is general and has been used in previous works where the Bhagavad Gita human expert translation has been compared \cite{chandra_2022_semantic}, along with the effectiveness of Google Translate \cite{shukla2023evaluation}. This study motivates future work to extend the framework for evaluating other languages, particularly high-resource languages, for effective evaluation as more resources and data are available, such as the case of European languages (such as Spanish, French, and Italian). This would better evaluate the cross-cultural strengths and limitations of LLMs in language translation. The framework can also be extended to study how certain languages influence and are coupled with each other, and how certain metaphors and linguistic expressions overlap. Examples include Urdu vs Hindi and Hindi vs Hinglish, which is an evolving language of the internet and prominent in social media platforms such as X (Twitter).  In future work, the framework can be extended to review cultural nuances and idiomatic expressions, requiring expert-labelled datasets (such as from Acharya et al. \cite{acharya2020towards})  for the languages under investigation. Furthermore, the framework can be extended to novel LLMs, including multimodal LLMs, so that real-time speech recognition for machine translation can be done. In summary, our study and framework have paved the way for the automated evaluation of language translation models.

\textcolor{black}{Finally, we stress that LLMs must translate sacred texts such as the Bhagavad Gita with caution. This is because the nature of the translations can vary and have different meanings, such as for sensitive topics, such as whether meat eating is prohibited in Hinduism. The subject of this discussion is through the translation of the term “tamasik” diet (Chapter 17, verses 8-10), which is often mistranslated as a meat-eating diet, while “tamas” translates to darkness. Therefore, some commentaries (such as interpretations by the International Society for Krishna Consciousness \cite{king2012krishna}) take it that the  Bhagavad Gita has prohibited meat.  Hence, the ethics around the impact of translations for various religious and sacred texts need to be taken into account, and the LLM translation needs to be verified by language experts.}



\section{Conclusion}\label{ccl}

We presented a framework for the evaluation of the translation of selected Indian languages, including Sanskrit (Bhagavad Gita), Telugu (Maha Prasthanam) and Hindi (Tamas) by  LLMs (GPT-3.5, GPT-4o, Gemini) and Google Translate. We used a combination of sentiment and semantic analysis to compare the translated text with translations by human experts.  Our findings suggest that while LLMs have made significant progress in translation accuracy, challenges remain in preserving sentiment and semantic integrity, especially in figurative and philosophical contexts.

The sentiment analysis revealed that GPT-based models gave the best performance at maintaining the sentiment polarity when compared to the human expert in the translated texts. The semantic analysis scores indicate that the LLMs provide better translation results when compared to Google Translate in terms of semantic richness and vocabulary.  We find that LLMs are generally better at translation in terms of sentiment and semantics when compared to Google Translate.  The variability in the models' performance in divergent text styles as well therefore, gives opportunities to tailor the models to suit the specific requirements of different languages. Furthermore, we found that LLMs furnish better translations even if they capture the context of the text. 

The results of this study could help in the development of more accurate and culturally sensitive translation systems for large language models. Prompting the LLMs with some additional context or any background information based on what text has been used for translation would help enhance the quality of translation, and can also be done. Future work can be done with the comparison of these models with other texts and languages with more data.


\subsection*{Code and Data}

The open source code and data can be found at  \footnote{\url{https://github.com/sydney-machine-learning/GPTvsGoogleTranslate}}.
 \bibliographystyle{ieeetr} 
 \bibliography{cas-refs}

\begin{thebibliography}{100}
\expandafter\ifx\csname url\endcsname\relax
  \def\url#1{\texttt{#1}}\fi
\expandafter\ifx\csname urlprefix\endcsname\relax\def\urlprefix{URL }\fi
\expandafter\ifx\csname href\endcsname\relax
  \def\href#1#2{#2} \def\path#1{#1}\fi

\bibitem{sergeinirenburg_1993_progress}
S.~Nirenburg, Progress in machine translation, Ios Press ; Tokyo, 1993.

\bibitem{cambria_2014_jumping}
E.~Cambria, B.~White, Jumping nlp curves: A review of natural language processing research [review article], IEEE Computational Intelligence Magazine 9 (2014) 48--57.
\newblock \href {https://doi.org/10.1109/mci.2014.2307227} {\path{doi:10.1109/mci.2014.2307227}}.

\bibitem{wang_2021_progress}
H.~Wang, H.~Wu, Z.~He, L.~Huang, K.~W. Church, Progress in machine translation, Engineering 18 (07 2021).
\newblock \href {https://doi.org/10.1016/j.eng.2021.03.023} {\path{doi:10.1016/j.eng.2021.03.023}}.

\bibitem{li_2017_deep}
H.~Li, \href{https://academic.oup.com/nsr/article/5/1/24/4107792}{Deep learning for natural language processing: advantages and challenges}, National Science Review 5 (2017) 24--26.
\newblock \href {https://doi.org/10.1093/nsr/nwx110} {\path{doi:10.1093/nsr/nwx110}}.
\newline\urlprefix\url{https://academic.oup.com/nsr/article/5/1/24/4107792}

\bibitem{wu_2019_deep}
S.~Wu, K.~Roberts, S.~Datta, J.~Du, Z.~Ji, Y.~Si, S.~Soni, Q.~Wang, Q.~Wei, Y.~Xiang, B.~Zhao, H.~Xu, \href{https://academic.oup.com/jamia/article-abstract/27/3/457/5651084}{Deep learning in clinical natural language processing: a methodical review}, Journal of the American Medical Informatics Association 27 (2019) 457--470.
\newblock \href {https://doi.org/10.1093/jamia/ocz200} {\path{doi:10.1093/jamia/ocz200}}.
\newline\urlprefix\url{https://academic.oup.com/jamia/article-abstract/27/3/457/5651084}

\bibitem{philippkoehn_2014_statistical}
P.~Koehn, Statistical machine translation, Cambridge University Press, 2014.

\bibitem{wu_2016_googles}
M.~Johnson, M.~Schuster, Q.~V. Le, M.~Krikun, Y.~Wu, Z.~Chen, N.~Thorat, F.~Vi{\'e}gas, M.~Wattenberg, G.~Corrado, et~al., Google’s multilingual neural machine translation system: Enabling zero-shot translation, Transactions of the Association for Computational Linguistics 5 (2017) 339--351.

\bibitem{bennett_1985_the}
W.~S. Bennett, J.~Slocum, The lrc machine translation system, Computational Linguistics 11 (1985) 111--121.
\newblock \href {https://doi.org/10.5555/1187874.1187877} {\path{doi:10.5555/1187874.1187877}}.

\bibitem{riveratrigueros_2021_machine}
I.~Rivera-Trigueros, Machine translation systems and quality assessment: a systematic review, Language Resources and Evaluation (04 2021).
\newblock \href {https://doi.org/10.1007/s10579-021-09537-5} {\path{doi:10.1007/s10579-021-09537-5}}.

\bibitem{yao_2023_empowering}
B.~Yao, M.~Jiang, D.~Yang, J.~Hu, \href{https://arxiv.org/abs/2305.14328}{Empowering llm-based machine translation with cultural awareness} (05 2023).
\newblock \href {https://doi.org/10.48550/arXiv.2305.14328} {\path{doi:10.48550/arXiv.2305.14328}}.
\newline\urlprefix\url{https://arxiv.org/abs/2305.14328}

\bibitem{zhao_2023_a}
W.~X. Zhao, K.~Zhou, J.~Li, T.~Tang, X.~Wang, Y.~Hou, Y.~Min, B.~Zhang, J.~Zhang, Z.~Dong, Y.~Du, C.~Yang, Y.~Chen, Z.~Chen, J.~Jiang, R.~Ren, Y.~Li, X.~Tang, Z.~Liu, P.~Liu, J.-Y. Nie, J.-R. Wen, A survey of large language models, arXiv (Cornell University) (03 2023).
\newblock \href {https://doi.org/10.48550/arxiv.2303.18223} {\path{doi:10.48550/arxiv.2303.18223}}.

\bibitem{chang_2024_a}
Y.~Chang, X.~Wang, J.~Wang, W.~Yuan, L.~Yang, K.~Zhu, H.~Chen, X.~Yi, C.~Wang, Y.~Wang, W.~Ye, Y.~Zhang, Y.~Chang, P.~S. Yu, Q.~Yang, X.~Xie, A survey on evaluation of large language models, ACM Transactions on Intelligent Systems and Technology 15 (01 2024).
\newblock \href {https://doi.org/10.1145/3641289} {\path{doi:10.1145/3641289}}.

\bibitem{min_2023_recent}
B.~Min, H.~Ross, E.~Sulem, B.~Veyseh, T.~H. Nguyen, O.~Sainz, E.~Agirre, I.~Heintz, D.~Roth, Recent advances in natural language processing via large pre-trained language models: A survey, ACM Computing Surveys 56 (06 2023).
\newblock \href {https://doi.org/10.1145/3605943} {\path{doi:10.1145/3605943}}.

\bibitem{aharoni_2019_massively}
R.~Aharoni, M.~Johnson, O.~Firat, Massively multilingual neural machine translation (01 2019).
\newblock \href {https://doi.org/10.18653/v1/n19-1388} {\path{doi:10.18653/v1/n19-1388}}.

\bibitem{veniaminveselovsky_2023_artificial}
V.~Veselovsky, M.~H. Ribeiro, R.~West, Artificial artificial artificial intelligence: Crowd workers widely use large language models for text production tasks, arXiv (Cornell University) (06 2023).
\newblock \href {https://doi.org/10.48550/arxiv.2306.07899} {\path{doi:10.48550/arxiv.2306.07899}}.

\bibitem{caliskan_2017_semantics}
A.~Caliskan, J.~J. Bryson, A.~Narayanan, Semantics derived automatically from language corpora contain human-like biases, Science 356 (2017) 183--186.
\newblock \href {https://doi.org/10.1126/science.aal4230} {\path{doi:10.1126/science.aal4230}}.

\bibitem{cahyawijaya_2024_llms}
S.~Cahyawijaya, H.~Lovenia, P.~Fung, Llms are few-shot in-context low-resource language learners, arXiv (Cornell University) (03 2024).
\newblock \href {https://doi.org/10.48550/arxiv.2403.16512} {\path{doi:10.48550/arxiv.2403.16512}}.

\bibitem{anilkumarsingh_2008_natural}
A.~K. Singh, Natural language processing for less privileged languages: Where do we come from? where are we going? (2008) 7--12.

\bibitem{cieri_2016_selection}
C.~Cieri, M.~Maxwell, S.~Strassel, J.~Tracey, Selection criteria for low resource language programs. (2016) 4543--4549.

\bibitem{tsvetkov_2017_opportunities}
Y.~Tsvetkov, Opportunities and challenges in working with low-resource languages (2017).

\bibitem{magueresse_lowresource}
A.~Magueresse, V.~Carles, E.~Heetderks, Low-resource languages: A review of past work and future challenges.

\bibitem{vaswani_2017_attention}
A.~Vaswani, N.~Shazeer, N.~Parmar, J.~Uszkoreit, L.~Jones, A.~N. Gomez, L.~Kaiser, I.~Polosukhin, \href{https://arxiv.org/abs/1706.03762}{Attention is all you need} (06 2017).
\newline\urlprefix\url{https://arxiv.org/abs/1706.03762}

\bibitem{schubert_2014_computational}
L.~Schubert, \href{https://plato.stanford.edu/entries/computational-linguistics/}{Computational linguistics (stanford encyclopedia of philosophy)} (2014).
\newline\urlprefix\url{https://plato.stanford.edu/entries/computational-linguistics/}

\bibitem{narang_2022_pathways}
S.~Narang, A.~Chowdhery, \href{https://research.google/blog/pathways-language-model-palm-scaling-to-540-billion-parameters-for-breakthrough-performance/}{Pathways language model (palm): Scaling to 540 billion parameters for breakthrou} (2022).
\newline\urlprefix\url{https://research.google/blog/pathways-language-model-palm-scaling-to-540-billion-parameters-for-breakthrough-performance/}

\bibitem{floridi_2020_gpt3}
L.~Floridi, M.~Chiriatti, Gpt-3: Its nature, scope, limits, and consequences, Minds and Machines 30 (2020) 681–694.

\bibitem{liu_2023_gpt}
X.~Liu, Y.~Zheng, Z.~Du, M.~Ding, Y.~Qian, Z.~Yang, J.~Tang, Gpt understands, too (2023).

\bibitem{kalyan_2023_a}
K.~S. Kalyan, A survey of gpt-3 family large language models including <span class="nocase">chatgpt and gpt-4</span> (2023).

\bibitem{openai_2022_introducing}
OpenAI, \href{https://openai.com/blog/chatgpt}{Introducing chatgpt} (11 2022).
\newline\urlprefix\url{https://openai.com/blog/chatgpt}

\bibitem{hendy_2023_how}
A.~Hendy, M.~Abdelrehim, A.~Sharaf, V.~Raunak, M.~Gabr, H.~Matsushita, Y.~J. Kim, M.~Afify, H.~H. Awadalla, \href{https://arxiv.org/abs/2302.09210}{How good are gpt models at machine translation? a comprehensive evaluation}, arXiv.org (02 2023).
\newblock \href {https://doi.org/10.48550/arXiv.2302.09210} {\path{doi:10.48550/arXiv.2302.09210}}.
\newline\urlprefix\url{https://arxiv.org/abs/2302.09210}

\bibitem{hacker_2023_regulating}
Regulating <span class="nocase">ChatGPT and other large generative AI</span> models.

\bibitem{liesenfeld_2023_opening}
Opening up ChatGPT: Tracking openness, transparency, and accountability in instruction-tuned text generators.

\bibitem{zhao_2024_explainability}
H.~Zhao, H.~Chen, F.~Yang, N.~Liu, H.~Deng, H.~Cai, S.~Wang, D.~Yin, M.~Du, Explainability for large language models: A survey, ACM Transactions on Intelligent Systems and Technology 15~(2) (2024) 1–38.

\bibitem{koco_2023_chatgpt}
J.~Kocoń, I.~Cichecki, O.~Kaszyca, M.~Kochanek, D.~Szydło, J.~Baran, J.~Bielaniewicz, M.~Gruza, A.~Janz, K.~Kanclerz, Chatgpt: Jack of all trades, master of none, Information Fusion 99 (2023) 101861.

\bibitem{wang_2019_learning}
Q.~Wang, B.~Li, T.~Xiao, J.~Zhu, C.~Li, D.~F. Wong, L.~S. Chao, Learning deep transformer models for machine translation (07 2019).
\newblock \href {https://doi.org/10.18653/v1/p19-1176} {\path{doi:10.18653/v1/p19-1176}}.

\bibitem{peng_2023_towards}
K.~Peng, L.~Ding, Q.~Zhong, L.~Shen, X.~Liu, M.~Zhang, Y.~Ouyang, D.~Tao, Towards making the most of chatgpt for machine translation (2023).

\bibitem{lee_2023_artificial}
T.~K. Lee, Artificial intelligence and posthumanist translation: Chatgpt versus the translator (2023).

\bibitem{chi_2020_crosslingual}
Z.~Chi, L.~Dong, F.~Wei, W.~Wang, X.-L. Mao, H.~Huang, Cross-lingual natural language generation via pre-training, Proceedings of the AAAI Conference on Artificial Intelligence 34 (2020) 7570--7577.
\newblock \href {https://doi.org/10.1609/aaai.v34i05.6256} {\path{doi:10.1609/aaai.v34i05.6256}}.

\bibitem{zhang_2023_sentiment}
W.~Zhang, Y.~Deng, B.~Liu, S.~J. Pan, L.~Bing, \href{https://arxiv.org/abs/2305.15005}{Sentiment analysis in the era of large language models: A reality check} (05 2023).
\newblock \href {https://doi.org/10.48550/arXiv.2305.15005} {\path{doi:10.48550/arXiv.2305.15005}}.
\newline\urlprefix\url{https://arxiv.org/abs/2305.15005}

\bibitem{chau_2021_specializing}
E.~C. Chau, N.~A. Smith, \href{https://arxiv.org/abs/2106.09063}{Specializing multilingual language models: An empirical study} (2021).
\newblock \href {https://doi.org/10.18653/v1/2021.mrl-1.5} {\path{doi:10.18653/v1/2021.mrl-1.5}}.
\newline\urlprefix\url{https://arxiv.org/abs/2106.09063}

\bibitem{ranathunga_neural}
S.~Ranathunga, E.-S. Lee, \href{https://arxiv.org/pdf/2106.15115.pdf}{Neural machine translation for low-resource languages: A survey}.
\newline\urlprefix\url{https://arxiv.org/pdf/2106.15115.pdf}

\bibitem{mohtashami_2023_learning}
A.~Mohtashami, M.~Verzetti, P.~K. Rubenstein, Learning translation quality evaluation on low resource languages from large language models, arXiv (Cornell University) (02 2023).
\newblock \href {https://doi.org/10.48550/arxiv.2302.03491} {\path{doi:10.48550/arxiv.2302.03491}}.

\bibitem{kumar_2022_annotated}
R.~Kumar, S.~Singh, S.~Ratan, M.~Raj, S.~Sinha, B.~Lahiri, V.~Seshadri, K.~Bali, A.~K. Ojha, \href{https://arxiv.org/abs/2206.12931}{Annotated speech corpus for low resource indian languages: Awadhi, bhojpuri, braj and magahi} (06 2022).
\newblock \href {https://doi.org/10.48550/arXiv.2206.12931} {\path{doi:10.48550/arXiv.2206.12931}}.
\newline\urlprefix\url{https://arxiv.org/abs/2206.12931}

\bibitem{sethi_2023_a}
N.~Sethi, A.~Dev, P.~Bansal, A novel neural machine translation approach for low-resource sanskrit-hindi language pair, ACM Transactions on Asian and Low-Resource Language Information Processing (04 2023).
\newblock \href {https://doi.org/10.1145/3591207} {\path{doi:10.1145/3591207}}.

\bibitem{soujanyaporia_2018_multimodal}
S.~Poria, A.~Hussain, E.~Cambria, S.~I.~P. Ag, Multimodal sentiment analysis, Cham, Switzerland Springer, 2018.

\bibitem{mustafa_2023_real}
H.~Mustafa, T.~Mansoor, S.~Yaqoob, S.~Mehmood, M.~M. Rahman, Real time personality analysis by tweet mining: Data mining, Journal of Innovative Computing and Emerging Technologies 3 (03 2023).
\newblock \href {https://doi.org/10.56536/jicet.v3i1.60} {\path{doi:10.56536/jicet.v3i1.60}}.

\bibitem{shukla_an}
A.~Shukla, C.~Bansal, S.~Badhe, M.~Ranjan, R.~Chandra, \href{https://arxiv.org/pdf/2303.07201.pdf}{An evaluation of google translate for sanskrit to english translation via sentiment and semantic analysis}.
\newline\urlprefix\url{https://arxiv.org/pdf/2303.07201.pdf}

\bibitem{goddard_2011_semantic}
C.~Goddard, Semantic analysis : a practical introduction, Oxford University Press, 2011.

\bibitem{malik_2021_multimodal}
S.~Malik, P.~Bansal, Multimodal semantic analysis with regularized semantic autoencoder, Journal of Intelligent and Fuzzy Systems (2021) 1--9\href {https://doi.org/10.3233/jifs-189759} {\path{doi:10.3233/jifs-189759}}.

\bibitem{krommyda_2020_semantic}
M.~Krommyda, V.~Kantere, Semantic analysis for conversational datasets: Improving their quality using semantic relationships, International Journal of Semantic Computing 14 (2020) 395--422.
\newblock \href {https://doi.org/10.1142/s1793351x2050004x} {\path{doi:10.1142/s1793351x2050004x}}.

\bibitem{post_2018_a}
M.~Post, \href{https://aclanthology.org/W18-6319/}{A call for clarity in reporting bleu scores} (10 2018).
\newblock \href {https://doi.org/10.18653/v1/W18-6319} {\path{doi:10.18653/v1/W18-6319}}.
\newline\urlprefix\url{https://aclanthology.org/W18-6319/}

\bibitem{gamon_2005_sentencelevel}
M.~Gamon, A.~Aue, M.~Smets, Sentence-level mt evaluation without reference translations: Beyond language modeling (05 2005).

\bibitem{mager_2020_GPTtoo}
M.~Mager, R.~F. Astudillo, T.~Naseem, A.~Sultan, Y.-S. Lee, R.~Florian, S.~Roukos, Gpt-too: A language-model-first approach for amr-to-text generation (01 2020).
\newblock \href {https://doi.org/10.18653/v1/2020.acl-main.167} {\path{doi:10.18653/v1/2020.acl-main.167}}.

\bibitem{zhang2023prompting}
B.~Zhang, B.~Haddow, A.~Birch, Prompting large language model for machine translation: A case study (2023).
\newblock \href {http://arxiv.org/abs/2301.07069} {\path{arXiv:2301.07069}}.

\bibitem{yang_2020_towards}
J.~Yang, M.~Wang, H.~Zhou, C.~Zhao, W.~Zhang, Y.~Yu, L.~Li, Towards making the most of bert in neural machine translation, Proceedings of the AAAI Conference on Artificial Intelligence 34 (2020) 9378--9385.
\newblock \href {https://doi.org/10.1609/aaai.v34i05.6479} {\path{doi:10.1609/aaai.v34i05.6479}}.

\bibitem{lo_2020_extended}
C.-k. Lo, Extended study on using pretrained language models and yisi-1 for machine translation evaluation. (2020) 895--902.

\bibitem{lo-2020-extended}
C.-k. Lo, \href{https://aclanthology.org/2020.wmt-1.99}{Extended study on using pretrained language models and {Y}i{S}i-1 for machine translation evaluation}, in: L.~Barrault, O.~Bojar, F.~Bougares, R.~Chatterjee, M.~R. Costa-juss{\`a}, C.~Federmann, M.~Fishel, A.~Fraser, Y.~Graham, P.~Guzman, B.~Haddow, M.~Huck, A.~J. Yepes, P.~Koehn, A.~Martins, M.~Morishita, C.~Monz, M.~Nagata, T.~Nakazawa, M.~Negri (Eds.), Proceedings of the Fifth Conference on Machine Translation, Association for Computational Linguistics, Online, 2020, pp. 895--902.
\newline\urlprefix\url{https://aclanthology.org/2020.wmt-1.99}

\bibitem{zhang2020bertscore}
T.~Zhang, V.~Kishore, F.~Wu, K.~Q. Weinberger, Y.~Artzi, Bertscore: Evaluating text generation with bert (2020).
\newblock \href {http://arxiv.org/abs/1904.09675} {\path{arXiv:1904.09675}}.

\bibitem{faheem_2024_improving}
M.~A. Faheem, K.~T. Wassif, H.~Bayomi, S.~M. Abdou, \href{https://www.nature.com/articles/s41598-023-51090-4#Sec18}{Improving neural machine translation for low resource languages through non-parallel corpora: a case study of egyptian dialect to modern standard arabic translation}, Scientific Reports 14 (2024) 2265.
\newblock \href {https://doi.org/10.1038/s41598-023-51090-4} {\path{doi:10.1038/s41598-023-51090-4}}.
\newline\urlprefix\url{https://www.nature.com/articles/s41598-023-51090-4#Sec18}

\bibitem{petroanu_2023_tracing}
D.-M. Petroșanu, A.~Pîrjan, A.~Tăbușcă, \href{https://www.mdpi.com/2079-9292/12/24/4957}{Tracing the influence of large language models across the most impactful scientific works}, Electronics 12 (2023) 4957.
\newblock \href {https://doi.org/10.3390/electronics12244957} {\path{doi:10.3390/electronics12244957}}.
\newline\urlprefix\url{https://www.mdpi.com/2079-9292/12/24/4957}

\bibitem{nasukawa_2003_sentiment}
T.~Nasukawa, J.~Yi, Sentiment analysis, Proceedings of the international conference on Knowledge capture - K-CAP '03 (2003).
\newblock \href {https://doi.org/10.1145/945645.945658} {\path{doi:10.1145/945645.945658}}.

\bibitem{liu_2012_a}
B.~Liu, L.~Zhang, A survey of opinion mining and sentiment analysis, Mining Text Data (2012) 415--463\href {https://doi.org/10.1007/978-1-4614-3223-4_13} {\path{doi:10.1007/978-1-4614-3223-4_13}}.

\bibitem{daudert_2021_exploiting}
T.~Daudert, Exploiting textual and relationship information for fine-grained financial sentiment analysis, Knowledge-Based Systems 230 (2021) 107389.
\newblock \href {https://doi.org/10.1016/j.knosys.2021.107389} {\path{doi:10.1016/j.knosys.2021.107389}}.

\bibitem{zhao_2024_a}
Z.~Zhao, L.~Alzubaidi, J.~Zhang, Y.~Duan, Y.~Gu, \href{https://www.sciencedirect.com/science/article/pii/S0957417423033092}{A comparison review of transfer learning and self-supervised learning: Definitions, applications, advantages and limitations}, Expert Systems with Applications 242 (2024) 122807.
\newblock \href {https://doi.org/10.1016/j.eswa.2023.122807} {\path{doi:10.1016/j.eswa.2023.122807}}.
\newline\urlprefix\url{https://www.sciencedirect.com/science/article/pii/S0957417423033092}

\bibitem{misra_2023_sarcasm}
R.~Misra, P.~Arora, Sarcasm detection using news headlines dataset, AI Open (02 2023).
\newblock \href {https://doi.org/10.1016/j.aiopen.2023.01.001} {\path{doi:10.1016/j.aiopen.2023.01.001}}.

\bibitem{chandra_2022_semantic}
R.~Chandra, V.~Kulkarni, Semantic and sentiment analysis of selected bhagavad gita translations using bert-based language framework, IEEE Access 10 (2022) 21291--21315.
\newblock \href {https://doi.org/10.1109/access.2022.3152266} {\path{doi:10.1109/access.2022.3152266}}.

\bibitem{nandwani_2021_a}
P.~Nandwani, R.~Verma, A review on sentiment analysis and emotion detection from text, Social Network Analysis and Mining 11 (08 2021).
\newblock \href {https://doi.org/10.1007/s13278-021-00776-6} {\path{doi:10.1007/s13278-021-00776-6}}.

\bibitem{medhat_2014_sentiment}
W.~Medhat, A.~Hassan, H.~Korashy, Sentiment analysis algorithms and applications: A survey, Ain Shams Engineering Journal 5 (2014) 1093--1113.
\newblock \href {https://doi.org/https://doi.org/10.1016/j.asej.2014.04.011} {\path{doi:https://doi.org/10.1016/j.asej.2014.04.011}}.

\bibitem{wilson_2009_recognizing}
T.~Wilson, J.~Wiebe, P.~Hoffmann, Recognizing contextual polarity: An exploration of features for phrase-level sentiment analysis, Computational Linguistics 35 (2009) 399--433.
\newblock \href {https://doi.org/10.1162/coli.08-012-r1-06-90} {\path{doi:10.1162/coli.08-012-r1-06-90}}.

\bibitem{park_2020_deep}
H.-j. Park, M.~Song, K.-S. Shin, Deep learning models and datasets for aspect term sentiment classification: Implementing holistic recurrent attention on target-dependent memories, Knowledge-Based Systems 187 (2020) 104825.
\newblock \href {https://doi.org/10.1016/j.knosys.2019.06.033} {\path{doi:10.1016/j.knosys.2019.06.033}}.

\bibitem{jaminrahmanjim_2024_recent}
J.~R. Jim, M.~Apon, P.~Malakar, M.~M. Kabir, K.~Nur, M.~Mridha, Recent advancements and challenges of nlp-based sentiment analysis: A state-of-the-art review, Natural Language Processing Journal (2024) 100059--100059\href {https://doi.org/10.1016/j.nlp.2024.100059} {\path{doi:10.1016/j.nlp.2024.100059}}.

\bibitem{taboada_2016_sentiment}
M.~Taboada, Sentiment analysis: An overview from linguistics, Annual Review of Linguistics 2 (2016) 325--347.
\newblock \href {https://doi.org/10.1146/annurev-linguistics-011415-040518} {\path{doi:10.1146/annurev-linguistics-011415-040518}}.

\bibitem{salloum_2020_a}
S.~A. Salloum, R.~Khan, K.~Shaalan, A survey of semantic analysis approaches, Advances in Intelligent Systems and Computing (2020) 61--70\href {https://doi.org/10.1007/978-3-030-44289-7_6} {\path{doi:10.1007/978-3-030-44289-7_6}}.

\bibitem{khurana_2022_natural}
D.~Khurana, A.~Koli, K.~Khatter, S.~Singh, Natural language processing: State of the art, current trends and challenges, Multimedia Tools and Applications 82 (2022) 3713–3744.
\newblock \href {https://doi.org/10.1007/s11042-022-13428-4} {\path{doi:10.1007/s11042-022-13428-4}}.

\bibitem{shukla2023evaluation}
A.~Shukla, C.~Bansal, S.~Badhe, M.~Ranjan, R.~Chandra, An evaluation of {Google Translate for Sanskrit to English} translation via sentiment and semantic analysis, Natural Language Processing Journal 4 (2023) 100025.

\bibitem{ddeepa_2021_bidirectional}
M.~D.Deepa, E.~Al, \href{https://www.turcomat.org/index.php/turkbilmat/article/view/3055}{Bidirectional encoder representations from transformers (bert) language model for sentiment analysis task: Review}, Turkish Journal of Computer and Mathematics Education (TURCOMAT) 12 (2021) 1708–1721.
\newblock \href {https://doi.org/10.17762/turcomat.v12i7.3055} {\path{doi:10.17762/turcomat.v12i7.3055}}.
\newline\urlprefix\url{https://www.turcomat.org/index.php/turkbilmat/article/view/3055}

\bibitem{lee_2021_combining}
S.~Lee, H.~Shin, Combining sentiment-combined model with pre-trained bert models for sentiment analysis, Journal of KIISE 48 (2021) 815--824.
\newblock \href {https://doi.org/10.5626/jok.2021.48.7.815} {\path{doi:10.5626/jok.2021.48.7.815}}.

\bibitem{devlin_2018_bert}
J.~Devlin, M.~Chang, K.~Lee, K.~Toutanova, Bert: Pre-training of deep bidirectional transformers for language understanding, arXiv (Cornell University) (10 2018).

\bibitem{ortizgarces_2024_optimizing}
I.~Ortiz-Garces, J.~Govea, R.~O. Andrade, W.~Villegas-Ch, \href{https://www.mdpi.com/2076-3417/14/5/1737}{Optimizing chatbot effectiveness through advanced syntactic analysis: A comprehensive study in natural language processing}, Applied Sciences 14 (2024) 1737.
\newblock \href {https://doi.org/10.3390/app14051737} {\path{doi:10.3390/app14051737}}.
\newline\urlprefix\url{https://www.mdpi.com/2076-3417/14/5/1737}

\bibitem{acheampong2021transformer}
F.~A. Acheampong, H.~Nunoo-Mensah, W.~Chen, Transformer models for text-based emotion detection: a review of {BERT-based} approaches, Artificial Intelligence Review 54~(8) (2021) 5789--5829.

\bibitem{song2020mpnet}
K.~Song, X.~Tan, T.~Qin, J.~Lu, T.-Y. Liu, {MPNet}: Masked and permuted pre-training for language understanding, Advances in neural information processing systems 33 (2020) 16857--16867.

\bibitem{caselli2020hatebert}
T.~Caselli, V.~Basile, J.~Mitrovi{\'c}, M.~Granitzer, {HateBERT} retraining {BERT} for abusive language detection in {English}, arXiv preprint arXiv:2010.12472 (2020).

\bibitem{zhang2020semantics}
Z.~Zhang, Y.~Wu, H.~Zhao, Z.~Li, S.~Zhang, X.~Zhou, X.~Zhou, Semantics-aware bert for language understanding, in: Proceedings of the AAAI conference on artificial intelligence, Vol.~34, 2020, pp. 9628--9635.

\bibitem{singh2025hp}
A.~Singh, R.~Chandra, {HP-BERT:} a framework for longitudinal study of {Hinduphobia on social media via LLMs}, arXiv preprint arXiv:2501.05482 (2025).

\bibitem{shripurohitswami_2010_the}
S.~P. Swami, The Bhagavad Gita, 2010.

\bibitem{prabhupada1972bhagavad}
A.~B.~S. Prabhupada, B.~Swami, Bhagavad-Gita as it is, Bhaktivedanta Book Trust Los Angeles, 1972.

\bibitem{jeste2008comparison}
D.~V. Jeste, I.~V. Vahia, Comparison of the conceptualization of wisdom in ancient {Indian literature with modern views: Focus on the Bhagavad Gita}, Psychiatry: Interpersonal and Biological Processes 71~(3) (2008) 197--209.

\bibitem{bayly2010india}
C.~A. Bayly, India, the {Bhagavad Gita} and the world, Modern Intellectual History 7~(2) (2010) 275--295.

\bibitem{nayak2018effective}
A.~K. Nayak, Effective leadership traits from {Bhagavad Gita}, International Journal of Indian Culture and Business Management 16~(1) (2018) 1--18.

\bibitem{bhatia2013bhagavad}
S.~C. Bhatia, J.~Madabushi, V.~Kolli, S.~K. Bhatia, V.~Madaan, The {Bhagavad Gita} and contemporary psychotherapies, Indian journal of psychiatry 55~(Suppl 2) (2013) S315--S321.

\bibitem{mukherjee2017bhagavad}
S.~Mukherjee, {Bhagavad Gita:} the key source of modern management, Asian J. Management 8~(1) (2017).

\bibitem{sastry1979sri}
K.~S. Sastry, {Sri Sri's' Mahaprasthanam':} a response, Indian Literature 22~(3) (1979) 56--61.

\bibitem{bhshmashan_2001_tamas}
B.~Sāhanī, Tamas, Penguin Group, 2001.

\bibitem{ahmed20021947}
I.~Ahmed, The 1947 partition of {India: A paradigm for pathological politics in India and Pakistan}, Asian ethnicity 3~(1) (2002) 9--28.

\bibitem{pandey2001remembering}
G.~Pandey, Remembering partition: Violence, nationalism, and history in {India}, Vol.~7, Cambridge University Press Cambridge, 2001.

\bibitem{daniels1996world}
P.~T. Daniels, W.~Bright, The world's writing systems, Oxford University Press, 1996.

\bibitem{google_2006_google}
Google, \href{https://translate.google.com}{Google translate} (04 2006).
\newline\urlprefix\url{https://translate.google.com}

\bibitem{johnson2017google}
M.~Johnson, M.~Schuster, Q.~V. Le, M.~Krikun, Y.~Wu, Z.~Chen, N.~Thorat, F.~Vi{\'e}gas, M.~Wattenberg, G.~Corrado, et~al., Google’s multilingual neural machine translation system: Enabling zero-shot translation, Transactions of the Association for Computational Linguistics 5 (2017) 339--351.

\bibitem{wu2016google}
Y.~Wu, Google's neural machine translation system: Bridging the gap between human and machine translation, arXiv preprint arXiv:1609.08144 (2016).

\bibitem{groves2015}
M.~Groves, K.~Mundt, Friend or foe? google translate in language for academic purposes, English for Specific Purposes 37 (2015) 112–121.
\newblock \href {https://doi.org/10.1016/j.esp.2014.09.001} {\path{doi:10.1016/j.esp.2014.09.001}}.

\bibitem{patil2014}
S.~Patil, P.~Davies, Use of google translate in medical communication: Evaluation of accuracy, BMJ: British medical journal 349 (2014) g7392.
\newblock \href {https://doi.org/10.1136/bmj.g7392} {\path{doi:10.1136/bmj.g7392}}.

\bibitem{Tsaishu2019}
S.-C. Tsai, Using google translate in efl drafts: a preliminary investigation, Computer Assisted Language Learning 32 (2019) 1--17.
\newblock \href {https://doi.org/10.1080/09588221.2018.1527361} {\path{doi:10.1080/09588221.2018.1527361}}.

\bibitem{Prates2020}
M.~Prates, P.~Avelar, L.~Lamb, Assessing gender bias in machine translation: a case study with google translate, Neural Computing and Applications 32 (05 2020).
\newblock \href {https://doi.org/10.1007/s00521-019-04144-6} {\path{doi:10.1007/s00521-019-04144-6}}.

\bibitem{wu2023brief}
T.~Wu, S.~He, J.~Liu, S.~Sun, K.~Liu, Q.-L. Han, Y.~Tang, A brief overview of {ChatGPT:} the history, status quo and potential future development, IEEE/CAA Journal of Automatica Sinica 10~(5) (2023) 1122--1136.

\bibitem{fui2023generative}
F.~Fui-Hoon~Nah, R.~Zheng, J.~Cai, K.~Siau, L.~Chen, Generative {AI and ChatGPT:} applications, challenges, and ai-human collaboration (2023).

\bibitem{rogers_2023_chatgpt}
R.~Rogers, \href{https://www.wired.com/story/chatgpt-vs-gemini-ai-chatbot-comparison/}{Chatgpt vs. gemini: Which ai chatbot subscription is right for you?} (2023).
\newline\urlprefix\url{https://www.wired.com/story/chatgpt-vs-gemini-ai-chatbot-comparison/}

\bibitem{labruna2023unraveling}
T.~Labruna, S.~Brenna, A.~Zaninello, B.~Magnini, Unraveling {ChatGPT}: A critical analysis of {AI}-generated goal-oriented dialogues and annotations, in: International Conference of the Italian Association for Artificial Intelligence, Springer, 2023, pp. 151--171.

\bibitem{lai_2023_chatgpt}
V.~D. Lai, N.~T. Ngo, B.~, H.~Man, F.~Dernoncourt, T.~Bui, T.~H. Nguyen, Chatgpt beyond english: Towards a comprehensive evaluation of large language models in multilingual learning (04 2023).
\newblock \href {https://doi.org/10.48550/arxiv.2304.05613} {\path{doi:10.48550/arxiv.2304.05613}}.

\bibitem{_2021_Gemini}
G.~, \href{https://deepmind.google/technologies/gemini/#introduction}{Gemini - google deepmind} (2021).
\newline\urlprefix\url{https://deepmind.google/technologies/gemini/#introduction}

\bibitem{syedaselinaakter_2023_an}
S.~S. Akter, Z.~Yu, A.~Muhamed, T.-Y. Ou, A.~Bäuerle, Ángel Alexander~Cabrera, K.~Dholakia, C.~Xiong, G.~Neubig, An in-depth look at gemini's language abilities, arXiv (Cornell University) (12 2023).
\newblock \href {https://doi.org/10.48550/arxiv.2312.11444} {\path{doi:10.48550/arxiv.2312.11444}}.

\bibitem{benneh2023}
G.~Benneh~Mensah, Artificial intelligence and ethics: A comprehensive review of bias mitigation, transparency, and accountability in ai systems (11 2023).
\newblock \href {https://doi.org/10.13140/RG.2.2.23381.19685/1} {\path{doi:10.13140/RG.2.2.23381.19685/1}}.

\bibitem{lee2023gemini}
G.-G. Lee, E.~Latif, L.~Shi, X.~Zhai, Gemini pro defeated by {GPT}-4v: Evidence from education, arXiv preprint arXiv:2401.08660 (2023).

\bibitem{borji2023}
A.~Borji, M.~Mohammadian, Battle of the wordsmiths: Comparing chatgpt, gpt-4, claude, and bard, SSRN Electronic Journal (06 2023).
\newblock \href {https://doi.org/10.2139/ssrn.4476855} {\path{doi:10.2139/ssrn.4476855}}.

\bibitem{ji_2020_leveraging}
Z.~Ji, L.~Dai, J.~Pang, T.~Shen, Leveraging concept-enhanced pre-training model and masked-entity language model for named entity disambiguation, IEEE Access (2020) 1--1\href {https://doi.org/10.1109/access.2020.2994247} {\path{doi:10.1109/access.2020.2994247}}.

\bibitem{yu_2019_masked}
G.~Yu, Z.~Zhang, H.~Liu, L.~Ding, Masked sentence model based on bert for move recognition in medical scientific abstracts, Journal of Data and Information Science 4 (2019) 42--55.
\newblock \href {https://doi.org/10.2478/jdis-2019-0020} {\path{doi:10.2478/jdis-2019-0020}}.

\bibitem{choi_2021_evaluation}
H.~Choi, J.~Kim, S.~Joe, Y.~Gwon, Evaluation of bert and albert sentence embedding performance on downstream nlp tasks (01 2021).
\newblock \href {https://doi.org/10.1109/ICPR48806.2021.9412102} {\path{doi:10.1109/ICPR48806.2021.9412102}}.

\bibitem{koloski_2022_out}
B.~Koloski, S.~Pollak, B.~Škrlj, M.~Martinc, Out of thin air: Is zero-shot cross-lingual keyword detection better than unsupervised?, arXiv (Cornell University) (01 2022).
\newblock \href {https://doi.org/10.48550/arxiv.2202.06650} {\path{doi:10.48550/arxiv.2202.06650}}.

\bibitem{nadim_2023_a}
M.~Nadim, D.~Akopian, A.~Matamoros, A comparative assessment of unsupervised keyword extraction tools, IEEE access 11 (2023) 144778--144798.
\newblock \href {https://doi.org/10.1109/access.2023.3344032} {\path{doi:10.1109/access.2023.3344032}}.

\bibitem{R2023}
S.~R, M.~Mujahid, F.~Rustam, R.~Shafique, V.~Chunduri, M.~G. Villar, J.~B. Ballester, I.~d. l.~T. Diez, I.~Ashraf, \href{http://dx.doi.org/10.3390/info14090474}{Analyzing sentiments regarding chatgpt using novel bert: A machine learning approach}, Information 14~(9) (2023) 474.
\newblock \href {https://doi.org/10.3390/info14090474} {\path{doi:10.3390/info14090474}}.
\newline\urlprefix\url{http://dx.doi.org/10.3390/info14090474}

\bibitem{liu_2019_roberta}
Y.~Liu, M.~Ott, N.~Goyal, J.~Du, M.~Joshi, D.~Chen, O.~Levy, M.~Lewis, L.~Zettlemoyer, V.~Stoyanov, \href{https://arxiv.org/abs/1907.11692}{Roberta: A robustly optimized bert pretraining approach} (07 2019).
\newline\urlprefix\url{https://arxiv.org/abs/1907.11692}

\bibitem{chandramouli2011census}
C.~Chandramouli, R.~General, Census of {India 2011}, Provisional Population Totals. New Delhi: Government of India (2011) 409--413.

\end{thebibliography}




\begin{IEEEbiography}[{\includegraphics[width=1in,height=1.25in,clip,keepaspectratio]{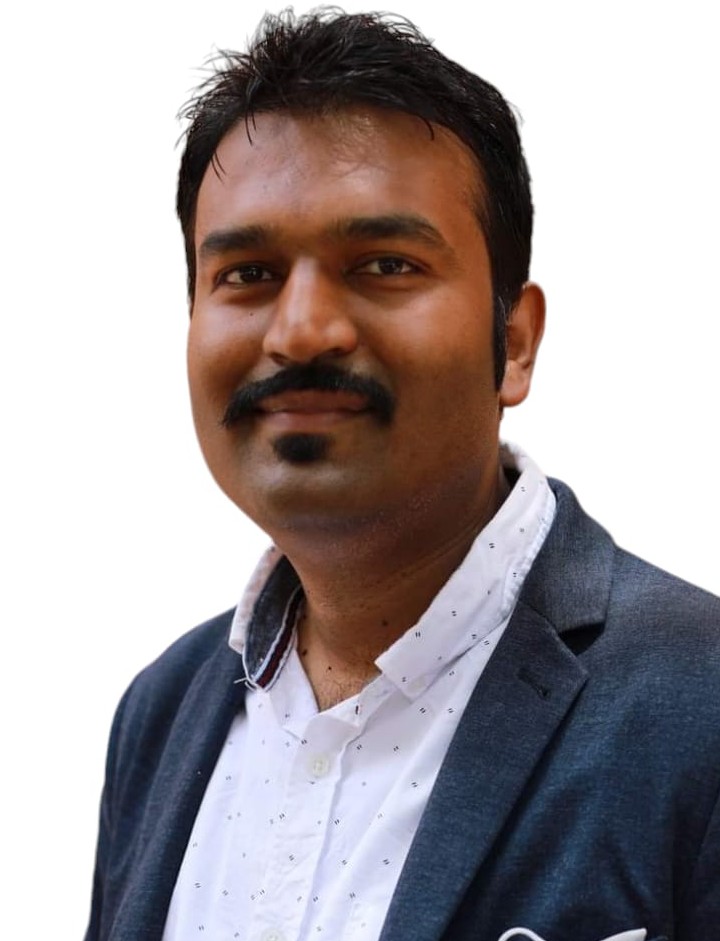}}]{Dr Rohitash Chandra}  is a Senior Lecturer in Data Science at the UNSW School of Mathematics and Statistics.  He leads a program of research encompassing methodologies and applications of artificial intelligence with a focus on Bayesian deep learning,  ensemble learning, and data augmentation.   Dr Chandra has pioneered the area of language models for studying ancient religious-philosophical texts and is currently focusing on the evaluation of large language models for problems in the arts and humanities.  

\end{IEEEbiography}

\begin{IEEEbiography}[{\includegraphics[width=1in,height=1.25in,clip,keepaspectratio]{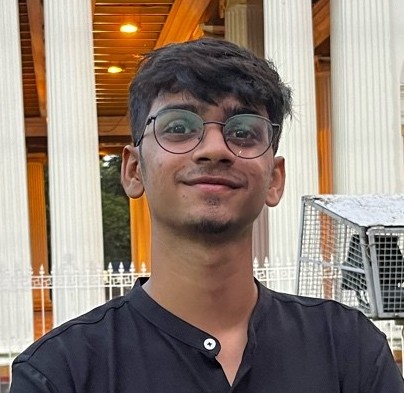}}]{Aryan Chaudhari}  is an undergraduate student in computer science at the National Institute of Technology, Rourkela, India. His research interests are in areas of natural language processing, deep learning and large language models. 

\end{IEEEbiography}

\begin{IEEEbiography}[{\includegraphics[width=1in,height=1.25in,clip,keepaspectratio]{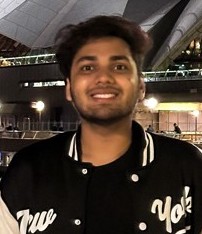}}]{Yeshwanth Rayavarapu}  graduated with a Master of Data Science at UNSW Sydney. His research interests are in areas of natural language processing, deep learning and large language models. 

\end{IEEEbiography}
\EOD

\end{document}